\documentclass{article}
\usepackage{etoolbox}
\usepackage{microtype}
\usepackage{graphicx}
\usepackage{subcaption}
\usepackage[table]{xcolor} % for table cell colors - must be before booktabs
\usepackage{booktabs} % for professional tables
\usepackage{makecell}
\usepackage[most]{tcolorbox}
\usepackage{xcolor}
\usepackage{dashrule}

\usepackage{hyperref}

\usepackage{booktabs}    
\usepackage{multirow}  
\usepackage{adjustbox} 
\usepackage{graphicx}

\newcommand{\bench}{AdaptMMBench}

\usepackage[preprint]{icml2026}

\usepackage{amsmath}
\usepackage{amssymb}
\usepackage{mathtools}
\usepackage{amsthm}

\usepackage[capitalize,noabbrev]{cleveref}

\theoremstyle{plain}

\theoremstyle{definition}

\theoremstyle{remark}

\usepackage[textsize=tiny]{todonotes}
\usepackage{marvosym} % for \Envelope symbol

\icmltitlerunning{\bench: Benchmarking Adaptive Multimodal Reasoning for Mode Selection and Reasoning Process}

\begin{document}

\twocolumn[
\icmltitle{\bench: Benchmarking Adaptive Multimodal Reasoning for Mode Selection and Reasoning Process}

\icmlsetsymbol{equal}{*}
\begin{icmlauthorlist}
\icmlauthor{Xintong Zhang$^{1,2,*}$}{}
\icmlauthor{Xiaowen Zhang$^{3,2,*}$}{}
\icmlauthor{Jingrong Wu$^{2,*}$}{}
\icmlauthor{Zhi Gao$^{1,2,4,\dagger,\clubsuit}$}{}
\icmlauthor{Shilin Yan$^{5,\dagger}$}{}
\icmlauthor{Zhenxin Diao$^{1,\ddagger}$}{}
\icmlauthor{Kunpeng Gao$^{1,\ddagger}$}{}
\icmlauthor{Xuanyan Chen$^{1,\ddagger}$}{}
\icmlauthor{Yuwei Wu$^{1,4,\clubsuit}$}{}
\icmlauthor{Yunde Jia$^{4}$}{}
\icmlauthor{Qing Li$^{2,\clubsuit}$}{}
\vspace{0.2cm}
\begin{tabular}{c}
\textsuperscript{1}Beijing Key Laboratory of Intelligent Information Technology, School of Computer Science \& Technology,\\Beijing Institute of Technology
\textsuperscript{2}State Key Laboratory of General Artificial Intelligence, BIGAI \textsuperscript{3}Xidian University\\
\textsuperscript{4}Guangdong Laboratory of Machine Perception and Intelligent Computing, Shenzhen MSU-BIT University \textsuperscript{5}Alibaba Group
\\
\vspace{0.3em}
$^{*}$ Core contribution, $^{\dagger}$ Project supervisor, $^{\ddagger}$ Equal contribution, $^{\clubsuit}$ Corresponding authors
   \end{tabular} 
% \vspace{0.3cm}
\\
Project Page: \url{https://adaptmmbench.github.io/}

\end{icmlauthorlist}

% \icmlaffiliation{sch}{Beijing Institute of Technology}
% \icmlaffiliation{bigai}{BIGAI}
% \icmlaffiliation{xidian}{Xidian University}
% \icmlaffiliation{ali}{Alibaba Group}
% \icmlaffiliation{msu}{Shenzhen MSU-BIT University}

\icmlcorrespondingauthor{Zhi Gao}{gaozhibit@bit.edu.cn}
\icmlcorrespondingauthor{Yuwei Wu}{wuyuwei@bit.edu.cn}
\icmlcorrespondingauthor{Qing Li}{dylan.liqing@gmail.com}
% \icmlcorrespondingauthor{Zhi Gao, Yuwei Wu, Qing Li}{\{gaozhibit, wuyuwei\}@bit.edu.cn, dylan.liqing@gmail.com}

\vskip 0.3in
]



\printAffiliationsAndNotice{}  % no special notice (required even if empty)
% \printAffiliationsAndNotice{$^*$Core contribution, $^\dagger$Project supervisor, $^\ddagger$Equal contribution.}

\begin{abstract}
  Adaptive multimodal reasoning has emerged as a promising frontier in Vision-Language Models (VLMs), aiming to dynamically modulate between tool-augmented visual reasoning and text reasoning to enhance both effectiveness and efficiency. However, existing evaluations rely on static difficulty labels and simplistic metrics, which fail to capture the dynamic nature of difficulty relative to varying model capacities. Consequently, they obscure the distinction between adaptive mode selection and general performance while neglecting fine-grained process analyses.
  In this paper, we propose \bench, a comprehensive benchmark for adaptive multimodal reasoning across five domains: real-world, OCR, GUI, knowledge, and math, encompassing both direct perception and complex reasoning tasks. \bench\ utilizes a Matthews Correlation Coefficient (MCC) metric to evaluate the selection rationality of different reasoning modes, isolating this meta-cognition ability by dynamically identifying task difficulties based on models' capability boundaries. Moreover, \bench\ facilitates multi-dimensional process evaluation across key step coverage, tool effectiveness, and computational efficiency. Our evaluation reveals that while adaptive mode selection scales with model capacity, it notably decouples from final accuracy. Conversely, key step coverage aligns with performance, though tool effectiveness remains highly inconsistent across model architectures.
\end{abstract}

\section{Introduction}

\begin{figure}[ht!]
  \centering
  \includegraphics[width=0.85\linewidth]{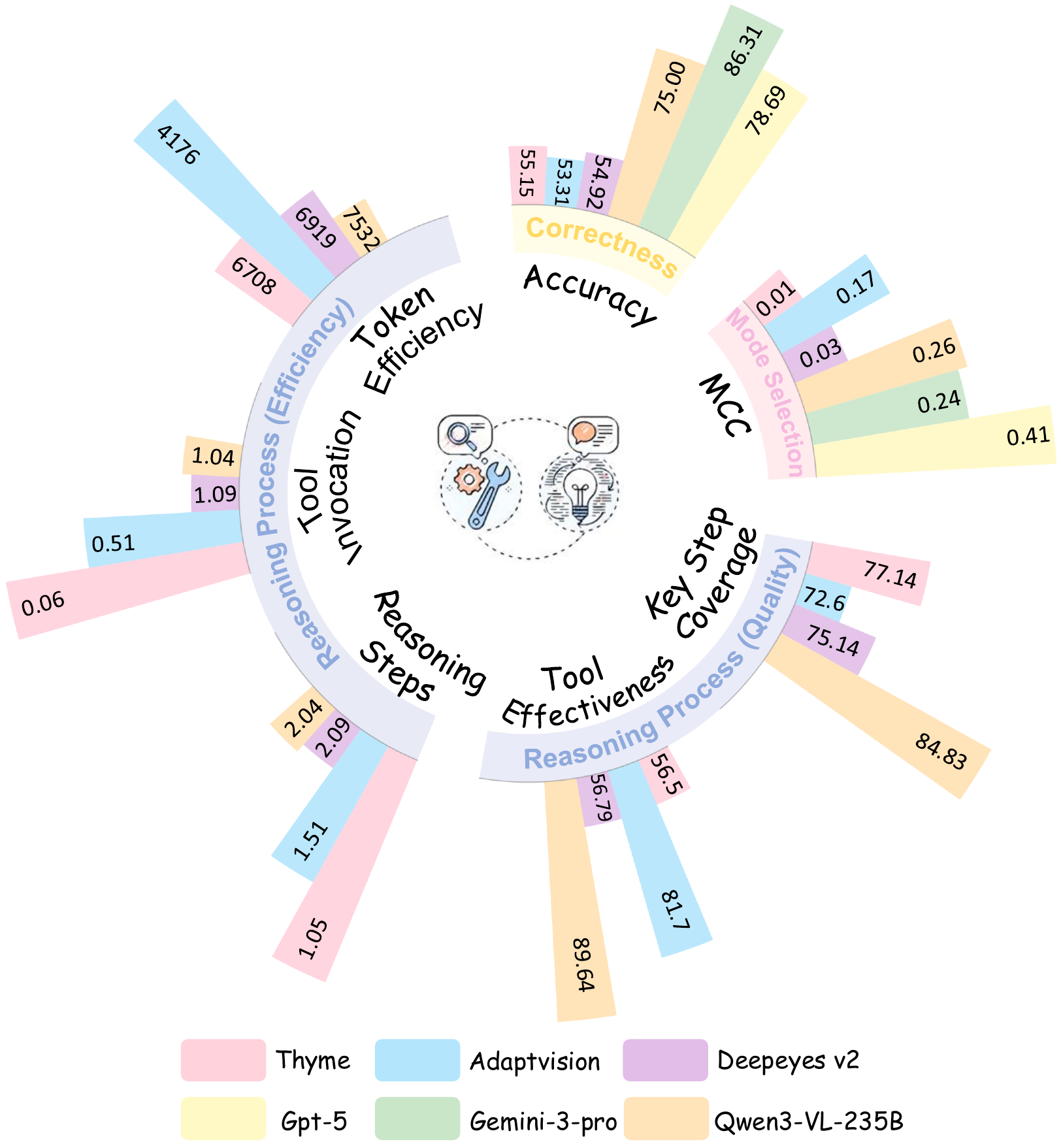}
  \caption{\textbf{Comparative Analysis of Accuracy, Reasoning Mode Selection, and Reasoning Process.} Closed-source models achieve stronger performance in accuracy and mode selection, while reasoning process quality is analyzed on open-source models due to limited access to closed-source reasoning traces.}
  \label{fig:performance}
\end{figure}

Vision Language Models (VLMs) have evolved from passive observers of static visual inputs to proactive models capable of dynamic information seeking.
This evolution makes a shift from direct perception and textual chain-of-thought (CoT) to the tool-augmented visual reasoning (i.e., thinking with images)~\cite{openai2025o3},
where models iteratively manipulate the visual content using visual tools, such as zoom-in, and enhancement(e.g., contrast and rotation) to acquire more visual information and resolve ambiguities.~\cite{zheng2025deepeyes,hu2024visual}.
However, this capability introduces significant computational redundancy. Lacking a mechanism to discern task necessity, models often fall into a `tool-invocation' trap, applying intensive visual tools to tasks solvable by direct perception or text reasoning. Consequently, \textbf{adaptive multimodal reasoning} is a promising direction for VLMs, which balances the necessity of such tool-augmented visual reasoning against text reasoning~\cite{lin2025adaptvision, wang2025adatooler}.

\begin{figure*}[ht!]
  \centering
  \includegraphics[width=0.70\linewidth]{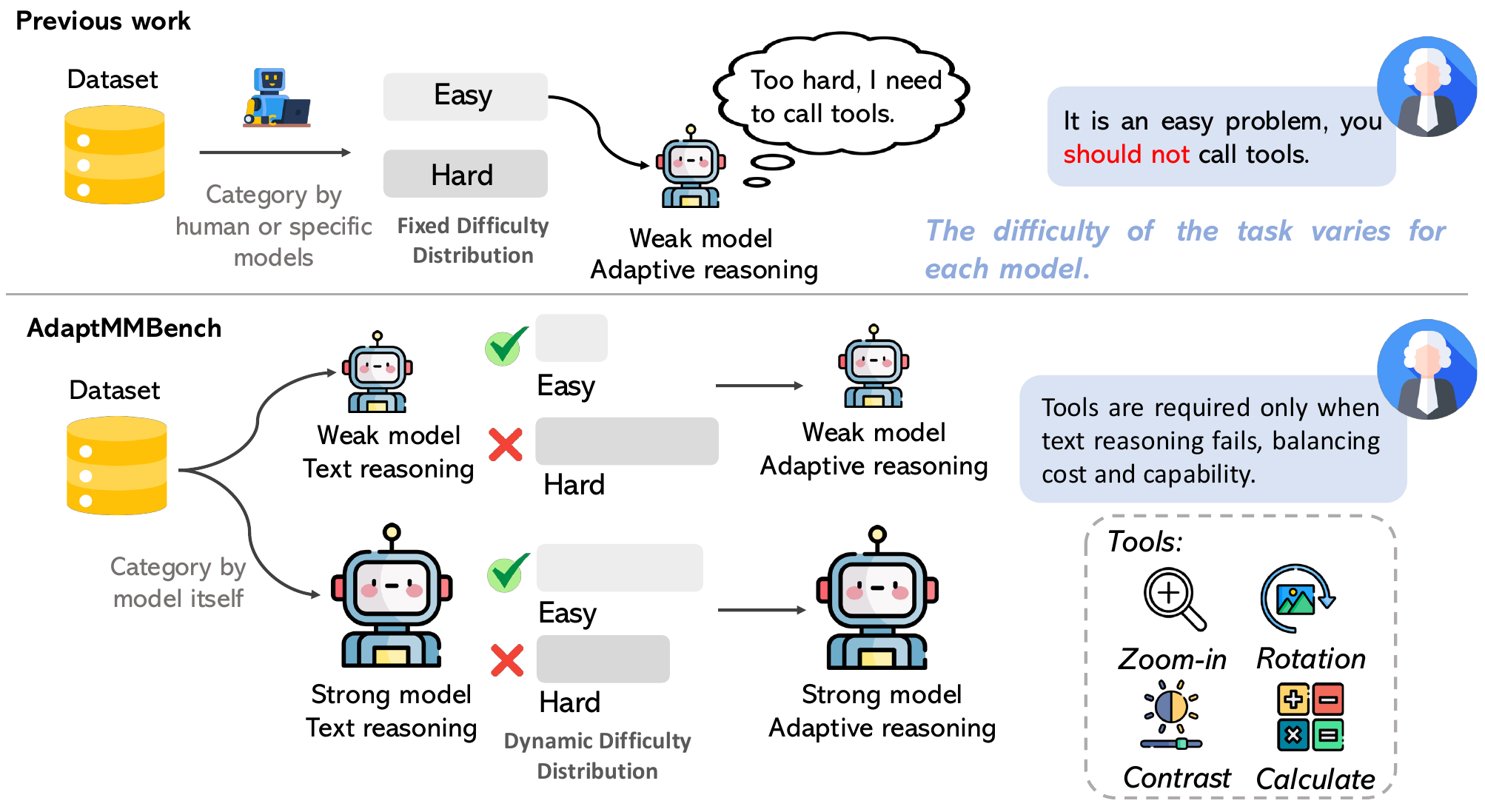}
  \caption{\textbf{Illustration of our model-specific difficulty evaluation.} Existing methods rely on static difficulty levels, while difficulty is inherently model-dependent.}
  \label{fig:teaser}
\end{figure*}

Despite the emergence of adaptive multimodal reasoning formulations, \textbf{evaluating adaptive multimodal reasoning remains an open problem.} Most existing evaluations rely on token-level reduction, coarse tool-call statistics, or final accuracy as proxies for adaptive intelligence. While intuitive, these metrics primarily reflect observable outcomes rather than evaluating the internal reasoning process itself. In particular, they fail to disentangle adaptive reasoning mode selection from subsequent reasoning execution. The ability to select an appropriate reasoning mode is crucial, as it reflects difficulty-aware meta-cognition.
From the data perspective, adaptive reasoning is commonly evaluated on domain-specific logic tasks (e.g., math and knowledge reasoning) or high-resolution perception benchmarks~\cite{wu2024v, wang2025divide, zhang2025mme}. These benchmarks lack a hierarchy of difficulty, limiting their effectiveness in evaluating adaptive reasoning. Recent efforts such as Omni-AutoThink~\cite{yang2025omni} attempt to quantify adaptiveness through thinking rates under predefined difficulty levels, as shown in Fig.~\ref{fig:teaser}. While this encourages increased reasoning effort on harder tasks, predefined difficulty levels are not universally applicable across models, leading to evaluation bias. Moreover, existing evaluations largely overlook reasoning process quality, losing detailed analyzes to guide future multimodal reasoning research.

To bridge these gaps, we propose AdaptMMBench to quantify adaptive multimodal reasoning in VLMs. AdaptMMBench includes 1420 samples across five domains: real-world, OCR, GUI, knowledge, and math. Each domain contains both text-only solvable tasks and complex scenarios of varying difficulties requiring proactive visual tool invocation. AdaptMMBench enables separate evaluation of adaptive reasoning mode selection and the reasoning process. Specifically, it adopts the Matthews Correlation Coefficient (MCC) to evaluate mode selection by dynamically identifying task difficulties based on model performance boundaries. For reasoning process evaluation, we assess key step coverage, tool invocation effectiveness, and efficiency to measure reasoning coherence, tool correctness, and computational cost alongside accuracy.

We evaluate closed-source and open-source VLMs on \bench, results shown in Fig.~\ref{fig:performance}. Experiments reveal a relatively weak correlation between adaptive mode selection performance and final accuracy, whereas closed-source and larger models demonstrate stronger adaptive capability. By contrast, key step coverage correlates more closely with accuracy, and tool execution effectiveness varies substantially across models.

Our contributions are summarized as follows.

(1) We propose the \bench\ to quantify the adaptive multimodal reasoning capabilities of VLMs, which contains 1420 samples across five domains with detailed reasoning annotations for comprehensive evaluations.

(2) We establish a suite of metrics for adaptive multimodal reasoning, which disentangle the adaptive capability from other model capabilities and assess three aspects of the reasoning process, providing detailed and in-depth evaluations.

(3) We analyze current VLMs from the perspective of adaptive reasoning, highlighting that the relationship between mode selection performance and final accuracy is relatively small, while closed-source and larger models exhibit stronger adaptive behavior. In contrast, key step coverage correlates more closely with accuracy, and tool execution effectiveness varies substantially across models.

\section{Related Work}

\subsection{Multimodal Reasoning in VLMs}

Early VLMs predominantly rely on text-only reasoning over fixed visual encodings, imposing a ``first-glance'' bottleneck that limits access to fine-grained visual details~\cite{lu2023mathvista, huang2025vision, zhang2023multimodal, yang2025magic}. Recent advanced models, including GPT-5~\cite{singh2025openai}, Qwen3-VL~\cite{Qwen3-VL}, and InternVL~\cite{zhu2025internvl3} have shifted multimodal reasoning from passive visual interpretation toward active, tool-augmented information seeking. Under this ``thinking with images'' paradigm, models acquire additional visual information through mechanisms such as multi-turn visual search~\cite{openai2025o3, zheng2025deepeyes}, region zoom-in~\cite{wang2025pixel, lai2025mini}, and self-generated visual cues~\cite{li2025imagine, chern2025thinking}. In parallel, adaptive multimodal reasoning models have emerged to selectively invoke tools, trading off between text-only and tool-based reasoning to improve inference efficiency~\cite{lin2025adaptvision, zhang2025chain, wang2025adatooler, li2025look, li2025mixture}. More advanced systems further incorporate agentic workflows and code generation to support precise execution~\cite{hong2025deepeyesv2, zhang2025thyme}. While these works emphasize improvements in precision and efficiency, they offer limited evaluation of whether models invoke tool-based reasoning until text-only reasoning is insufficient, avoiding unnecessary computational overhead.

\subsection{Benchmarks for VLMs}
Traditional VLM benchmarks mainly assess multimodal reasoning in structured domains with coarse visual content, such as chart understanding~\cite{mathew2021docvqa, masry2022chartqa}, mathematical problem solving~\cite{lu2023mathvista, xiao2024logicvista}, and other general-purpose VQA~\cite{liu2024mmbench, chen2024we}. MME-CoT~\cite{jiang2025mme} further evaluates the correctness of the text reasoning process. 
As VLM capabilities improve, more recent benchmarks~\cite{wu2024v, wang2025divide} introduce higher-resolution images to better reflect complex conditions. Building on this trend, benchmarks such as VisualProbe~\cite{lai2025mini}, InSight-o3~\cite{li2025insight}, and TIR-Bench~\cite{li2025tir} emphasize fine-grained visual understanding and active visual reasoning through operations like region zoom-in and iterative exploration, implicitly requiring models to ``thinking with images''. In parallel, generative benchmarks including VTBench~\cite{lin2025vtbench} and AuxSolidMath~\cite{guo2025geovlmath} evaluate multimodal reasoning via self-produced auxiliary visual cues, extending visual reasoning beyond the information directly available in the input image. However, these visual-grounded benchmarks~\cite{li2025insight,lin2025vtbench} largely focus on task accuracy, overlooking the problem of redundant computation where models use visual tools for tasks already solvable through text-only reasoning. Relying solely on token reduction for efficiency evaluation fails to evaluate the adaptive decisions and the reasoning quality.

\begin{figure*}[ht!]
  \centering
  \includegraphics[width=0.9\linewidth]{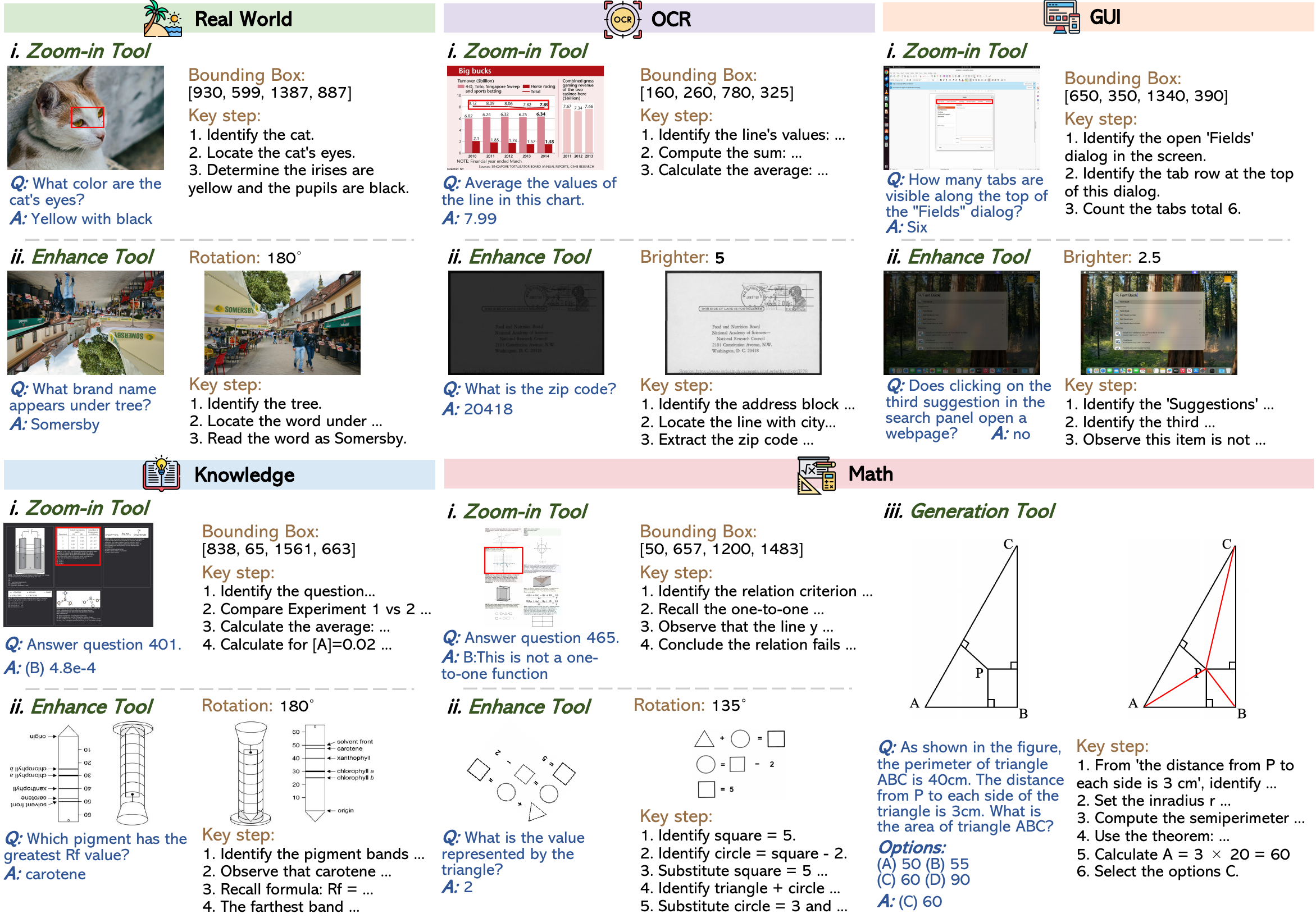}
  \caption{\textbf{An Overview of \bench.} The benchmark contains data from five domains. Each domain includes samples requiring zoom-in and enhancement tools. We annotate zoom-in regions, enhancement arguments, and key reasoning steps.}
  \label{fig:bench_domains}
\end{figure*}

\begin{figure}[ht!]
  \centering
  \includegraphics[width=0.75\linewidth]{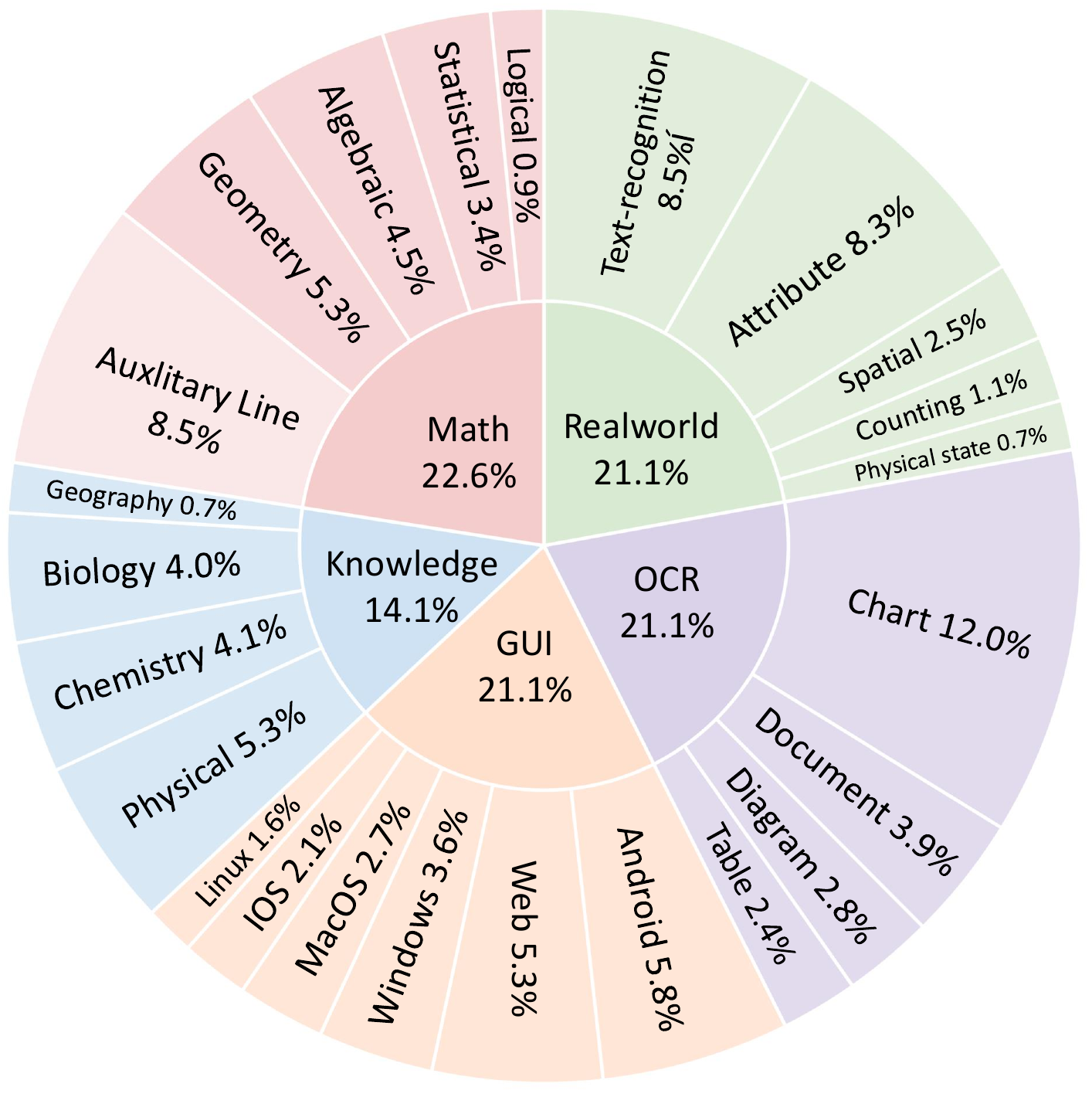}
  \caption{Domains and category of \bench.}
  \label{fig:data_distribution_pie}
\end{figure}

\section{\bench}

\bench\ focuses on two perspectives: adaptive reasoning mode selection and reasoning process.

\subsection{Data Formulation}
\label{sec:data_formulation}
Formally, \bench\ is constructed as a set of samples $\mathcal{D} = \{d_i\}_{i=1}^{N}$, where each data sample $d_i$ is defined as:
\begin{equation}
  d_i = (I, Q, A, E, K).
\end{equation}
Here, $I \in \mathbb{R}^{H \times W \times 3}$ denotes the input image, $Q$ is the textual query, and $A$ is the ground-truth answer. To support adaptive evaluation, we provide the visual tool annotation $E$ that specifies how essential visual information can be obtained, including the coordinates of target regions as well as the required image transformations such as rotation and contrast adjustment. $K = \{k_1, \dots, k_m\}$ is an ordered sequence of human-verified key reasoning steps describing the solution path from $(I,Q)$ to $A$.

During inference, the model only observes the image $I$ and the query $Q$. Acquiring the visual information specified by $E$ requires invoking a visual tool $t(I,\tau)$ via code execution or function calls, where $t \in \mathcal{T}$ denotes a tool from the predefined toolset and $\tau$ its execution arguments.

\subsection{Data Collection}

\bench\ encompasses 1,420 samples spanning five domains: real-world, OCR, GUI, math, and knowledge, enabling a comprehensive evaluation of adaptive reasoning across diverse scenarios, as detailed in Fig.~\ref{fig:bench_domains}.

To ensure that \bench\ contains both samples solvable via text-only reasoning and samples that require visual tool invocation under adaptive reasoning, we deliberately construct the dataset with diverse difficulty levels during data collection. One subset consists of samples solved by Qwen2.5-VL-7B under text-only reasoning. A second subset includes samples that Qwen2.5-VL-7B fails but can be solved by Qwen3-VL-235B based on adaptive reasoning. A small portion remains unsolved even by Qwen3-VL-235B. The relative proportions of these three subsets are approximately 24\%, 70\%, and 6\%. Notably, these subsets are introduced only to ensure difficulty diversity and do not determine the ground truth reasoning mode during evaluation. The reasoning mode selection label in adaptive mode is made by the model itself, as detailed in Sec.~\ref{sec:metrics}.

Building on prior adaptive reasoning methods~\cite{chern2025thinking, zhang2025thyme, zhao2025pyvision}, \bench\ evaluates diverse visual tools beyond zoom-in, including geometric transformations for orientation correction and photometric adjustments for visual enhancement. During data construction, these requirements are induced via controlled distortions such as changes in contrast, brightness, and orientation, with zoom-in and transformation samples with a ratio of 5:2. We further include 120 samples requiring auxiliary-line generation, suggesting that reasoning with self-generated images constitutes an important extension of the think-with-images paradigm.

\begin{figure*}[ht!]
  \centering
  \includegraphics[width=0.8\linewidth]{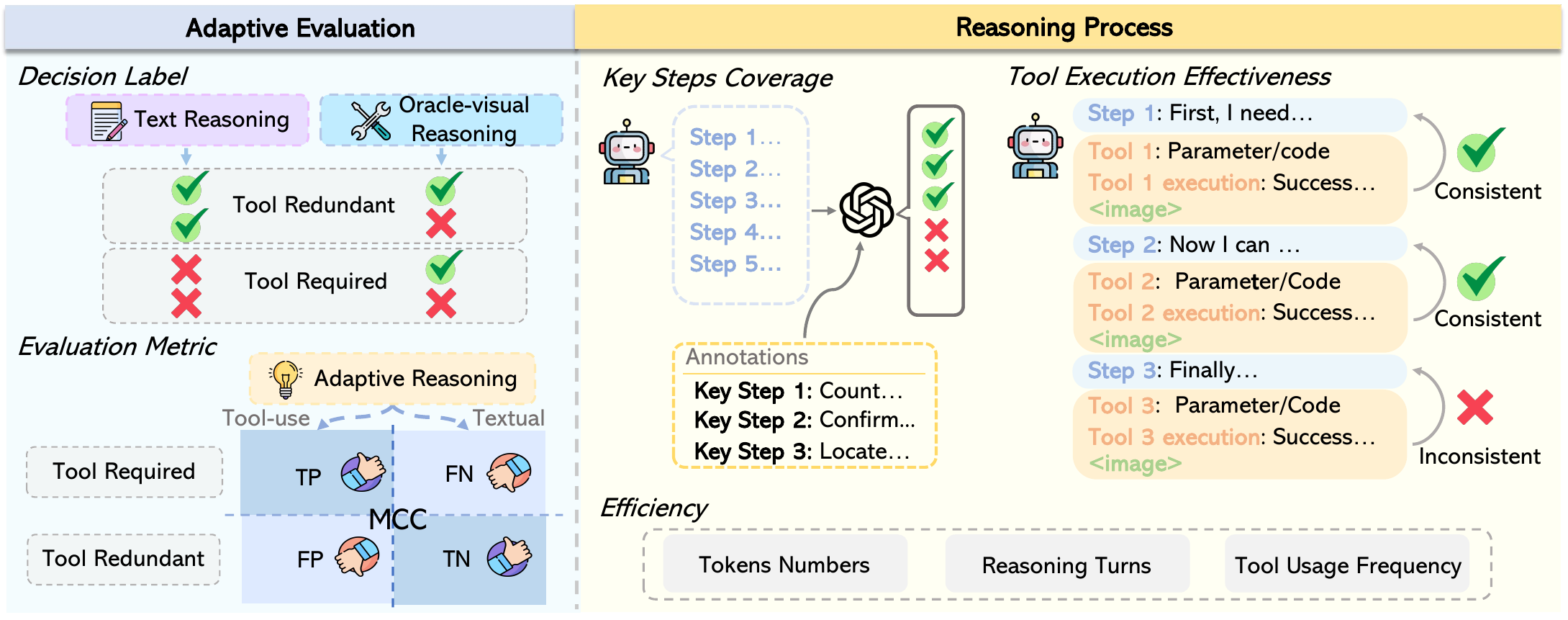}
  \caption{Evaluation pipeline for mode selection and reasoning process.}
  \label{fig:adaptive_evaluation}
\end{figure*}

\subsection{Annotation and Quality Control}

\noindent\textbf{Visual Tool \& Key Step Annotation.} 

We collect initial data from existing benchmarks, with annotators providing bounding-box annotations for key regions, while visual enhancement annotations are generated through predefined transformations. Distortion parameters are constrained to maintain recoverability. GPT-5 is used to generate key reasoning steps $K$, which are manually verified. These components form annotated quintuples $(I, Q, A, E, K)$.

\noindent\textbf{Quality Control.}
Benchmark quality is ensured through a multi-stage verification pipeline. First, three independent annotators cross-validate each QA pair to remove ambiguity and verify correctness. Annotated image transformations and generated key reasoning steps are then reviewed by additional annotators for precision. Inaccurate instances are iteratively refined or re-annotated. This process ensures high-fidelity ground truth with precise pixel-level annotations and reliable key reasoning steps for comprehensively evaluating adaptive reasoning. More statistical information of \bench\ can be found in Appendix~\ref{sec:Data_Source_Distribution}.

\section{Evaluation Strategy}
\label{sec:metrics}

\subsection{Evaluation Modes}

Following the formulation defined in Sec.~\ref{sec:data_formulation}, we define three evaluation modes to systematically assess the model's adaptive reasoning capabilities.
\begin{itemize}
  \item \textbf{Text-Reasoning Mode:} Given $(I, Q)$, the model relies solely on text reasoning over the given image, without invoking active visual transformations, providing a baseline for assessing tool necessity.
  \item \textbf{Adaptive Reasoning Mode:} Given $(I, Q)$, the model adaptively selects between text-only reasoning and tool-augmented visual reasoning with tools. It generates a reasoning trajectory and records all tool invocation parameters, enabling evaluation of both its ability to decide when tool usage is required and the correctness of the reasoning process.
  \item \textbf{Oracle-Visual Mode:} Given $(I, Q, I_{E})$, where $I_E$ denotes gold-standard visual evidence from annotation $E$, the model performs text-only reasoning over the provided visual evidence, providing an upper-bound performance estimate under perfect visual acquisition.
\end{itemize}

\subsection{Adaptive Mode Selection Evaluation}
Adaptive intelligence depends on a model’s ability to assess whether its current information is sufficient to solve a task. Consequently, the appropriateness of a reasoning mode should be evaluated independently of answer correctness.

Under this principle, the necessity of tool invocation is determined by the outcome of text-only reasoning. 
If a task can be solved using text reasoning alone, it is labeled as \textbf{Tool-Redundant}, indicating that visual tool invocation is unnecessary and may introduce noise. Conversely, tasks that cannot be solved via text-only reasoning are labeled as \textbf{Tool-Required}, indicating that visual tool invocation is necessary to obtain additional information. This categorization defines the mode selection labels used in our evaluation, as detailed in Fig.~\ref{fig:adaptive_evaluation}. Accordingly, tool invocation decisions are evaluated using a confusion matrix: TP denotes Tool-Required cases where the model invokes tools, FN denotes Tool-Required cases where the model does not invoke tools, TN denotes Tool-Redundant cases where the model selects text-only reasoning, and FP denotes Tool-Redundant cases where the model unnecessarily invokes tools.

\paragraph{Matthews Correlation Coefficient (MCC).}
In adaptive mode selection, the proportions of tool-redundant and tool-required cases are model-dependent, leading to varying degrees of class imbalance in the resulting confusion matrix. To ensure a robust evaluation, we adopt the MCC,
\begin{equation}
  \resizebox{0.9\columnwidth}{!}{$\displaystyle
      \text{MCC} =
      \frac{TP \cdot TN - FP \cdot FN}
      {\sqrt{(TP+FP)(TP+FN)(TN+FP)(TN+FN)} + \epsilon},
    $}
  \label{equal:mcc}
\end{equation}
where $\epsilon$ is a small constant for numerical stability. MCC ranges from $[-1,1]$, with $1$ indicating perfect agreement with the optimal mode selection, $0$ denoting the chance-level performance, and $-1$ indicating complete misalignment.

\paragraph{Adaptive Label Robustness.}
We analyze the effects of minor prompt variations on text and adaptive reasoning. Only 0.02 of samples show inconsistent outcomes between text reasoning mode and text-only reasoning in adaptive mode. This indicates that the performance difference is stable under prompt variations, and adaptive reasoning rarely degrades text-solvable samples.

\subsection{Reasoning Process Evaluation}

While MCC measures the quality of mode selection, it does not assess the validity of the reasoning process. Models may produce correct answers despite logical errors or improper tool usage. To address this limitation, we introduce three process-oriented metrics to evaluate reasoning coherence and tool execution fidelity.

A reasoning trajectory $\mathcal{R}$ is formalized as an interleaved sequence of reasoning steps and tool invocations:
\begin{equation}
  \mathcal{R} = \{(s_1, t_1), (s_2, t_2), \dots, s_n\},
\end{equation}
where $s_i$ is the reasoning at step $i$ and $t_i \in \mathcal{T}$ represents the corresponding tool invocation. The trajectory terminates at the final reasoning step $s_n$, and produces the answer.

\subsubsection{Key Steps Coverage}
Following the evaluation paradigm of \cite{jiang2025mme},
we assess whether a model’s reasoning chain $\{s_i\}_{i=1}^n$ covers the essential human-annotated key steps $K$ defined in Sec.~\ref{sec:data_formulation}. We employ GPT-5 as an evaluator to identify the presence of these key steps within the generated reasoning, and define the key step coverage as:
\begin{equation}
  \text{KCoverage}
  =
  \frac{1}{|K|}
  \max_j
  \prod_{i=1}^j
  \mathbb{I}\!\left[k_i \underset{\text{\tiny GPT-5}}{\in} \{s_1,\dots,s_n\}\right].
\end{equation}
This metric measures how far the model’s reasoning progresses along the key steps. Rather than penalizing skipped or compressed steps, KCoverage captures the maximum extent to which the reasoning aligns with the solution structure, allowing different reasoning styles and reflecting how close the model comes to a correct solution.

\subsubsection{Tool Execution Effectiveness}
To assess the precision of tool usage, we evaluate whether each tool invocation is semantically appropriate for its corresponding reasoning step and free of execution errors. The tool effectiveness is defined as:
\begin{equation}
  \text{Effect}_{\text{tool}}
  =
  \frac{1}{N_{\text{tool}}}
  \sum_{i=1}^{N_{\text{tool}}}
  \text{valid}_{\text{GPT-5}}(t_i \mid s_i),
\end{equation}
where $N_{\text{tool}}$ denotes the total number of tool invocations, $t_i \in \mathcal{T}$ is the tool invoked at step $i$, and $\text{valid}_{\text{GPT-5}}(\cdot) \in \{0,1\}$ is a semantic validity judgment provided by GPT-5.

\subsubsection{Reasoning Efficiency}
Efficiency is evaluated in terms of token numbers, reasoning turns, and tool usage frequency, collectively capturing the conciseness of reasoning and the computational cost of adaptive execution.

\begin{table}[ht!]
  \centering
  \definecolor{softblue}{rgb}{0.93, 0.96, 1.0}
  \definecolor{softgreen}{rgb}{0.94, 0.98, 0.94}
  \definecolor{reasoningyellow}{rgb}{1.0, 0.98, 0.9}
  \definecolor{opensourcecyan}{rgb}{0.9, 0.98, 1.0}
  \definecolor{closedpurple}{rgb}{0.96, 0.94, 1.0}
  \definecolor{headergray}{rgb}{0.95, 0.95, 0.95}
  \definecolor{darkergray}{rgb}{0.35, 0.35, 0.35}
  \caption{{Evaluation of mode selection performance across models.} We report TP, FP, TN, FN, and the MCC to assess meta-cognitive calibration in adaptive reasoning mode. Best and second-best scores in each category are highlighted in \colorbox{softblue}{blue} and \colorbox{softgreen}{green}.}
  \label{tab:adaptive_results}
  \begin{adjustbox}{width=0.4\textwidth}
    \begin{tabular}{l | cccc | c}
      \toprule
      \textbf{Model}         & \textbf{TP} & \textbf{FP} & \textbf{TN} & \textbf{FN}  & \textbf{MCC} $\uparrow$           \\
      \midrule
      \rowcolor{headergray} \multicolumn{6}{c}{\textit{Open-Source Models}} \\
      \midrule
      PixelReasoner          & 280                    & 196                      & 434                    & 390                      & 0.11                              \\
      Deepeyes               & 662                    & 638                      & 0                      & 0                        & 0.00                              \\
      Thyme                  & 20                     & 20                       & 655                    & 605                      & 0.01                              \\
      PyVision               & 540                    & 405                      & 231                    & 124                      & 0.20                              \\
      Deepeyes v2           & 623                    & 676                      & 1                      & 0                        & 0.03                              \\
      AdaptVision            & 385                    & 279                      & 375                    & 261                      & 0.17                              \\
      Qwen3-vl-8B-Instruct   & 328                    & 381                      & 351                    & 240                      & 0.06                              \\
      Qwen3-vl-32B-Instruct  & 348                    & 646                      & 245                    & 61                       & 0.14                              \\
      Qwen3-vl-235B-Instruct & 286                    & 437                      & 487                    & 90                       & \cellcolor{softgreen}0.26         \\
      \midrule
      \rowcolor{headergray} \multicolumn{6}{c}{\textit{Closed-Source Models}}   \\         
      \midrule
      GPT-5                  & 482                    & 392                      & 376                    & 50                       & \cellcolor{softblue}\textbf{0.41} \\
      Gemini-3-Pro           & 284                    & 703                      & 296                    & 17                       & 0.24                              \\
      \bottomrule
    \end{tabular}
  \end{adjustbox}
\end{table}

\begin{table}[h]
  \centering
  \definecolor{softblue}{rgb}{0.93, 0.96, 1.0}
  \definecolor{softgreen}{rgb}{0.94, 0.98, 0.94}
  \definecolor{headergray}{rgb}{0.95, 0.95, 0.95}
  \definecolor{darkergray}{rgb}{0.35, 0.35, 0.35}

  \caption{Comprehensive evaluation of reasoning process, including key step coverage (Key Step Cov.), tool effectiveness (Tool Effect.), and efficiency. This assesses the logical rigor of the reasoning paths alongside their computational efficiency.}
  \label{tab:process_results}
  \begin{adjustbox}{width=0.46\textwidth}
    \begin{tabular}{l cc | ccc}
      \toprule
      \multirow{2}{*}{\textbf{Model}} & \multirow{2}{*}{\shortstack{\textbf{Key Step} \\ \textbf{Cov.} (\%) $\uparrow$}} & \multirow{2}{*}{\shortstack{\textbf{Tool} \\ \textbf{Effect.} (\%) $\uparrow$}} & \multicolumn{3}{c}{\textbf{Efficiency}} \\ % 去掉了右侧竖线和 AEI
      \cmidrule(lr){4-6}
                                      &                                               &                    & Steps $\downarrow$ & Tools $\downarrow$ & Tokens $\downarrow$ \\
      \midrule
      PixelReasoner                   & 76.02                                         & 56.51              & \cellcolor{softgreen}{1.37}               & \cellcolor{softgreen}0.37                & 4229.00             \\
      Deepeyes                        & 75.56                                         & 50.99              & 2.00                & 1.68                & 7601.45             \\
      Thyme                           & 77.14                                         & 56.50              & \cellcolor{softblue}\textbf{{1.05}}   & \cellcolor{softblue}\textbf{0.06}      & 6708.47             \\
      PyVision                        & 77.43                                         & 62.02              & 2.76                & 1.76                & \cellcolor{softblue}\textbf{2481.00} \\
      Deepeyes v2                    & 75.14                                         & 56.79              & 2.09                & 1.09                & 6918.90             \\
      AdaptVision                     & 72.60                                         & 81.70              & 1.51                & 0.51   & \cellcolor{softgreen}4175.96             \\
      Qwen3-vl-8B                     & 78.40                                         & \cellcolor{softgreen}{91.62} & 1.76                & 1.20                & 8282.40             \\
      Qwen3-vl-32B                    & \cellcolor{softgreen}83.79                                         & \cellcolor{softblue}\textbf{92.98}    & 2.42                & 1.44                & 7725.99             \\
      Qwen3-vl-235B                   & \cellcolor{softblue}\textbf{84.83}                             & 89.64              & 2.04                & 1.04                & 7531.95             \\
      \bottomrule
    \end{tabular}
  \end{adjustbox}
\end{table}

\section{Experiments}
We conduct a comprehensive quantitative evaluation of adaptive reasoning on \bench, focusing on three complementary dimensions: (i) reasoning mode selection capability, (ii) quality and efficiency of reasoning process, and (iii) accuracy across reasoning modes.

\begin{table*}[ht!]
  \centering
  \definecolor{softblue}{rgb}{0.93, 0.96, 1.0}
  \definecolor{softgreen}{rgb}{0.94, 0.98, 0.94}
  \definecolor{reasoningyellow}{rgb}{1.0, 0.98, 0.9}
  \definecolor{opensourcecyan}{rgb}{0.9, 0.98, 1.0}
  \definecolor{closedpurple}{rgb}{0.96, 0.94, 1.0}
  \definecolor{headergray}{rgb}{0.95, 0.95, 0.95}
  \definecolor{darkergray}{rgb}{0.35, 0.35, 0.35}
  \caption{{Accuracy across different domains under three reasoning modes.} Results are reported on 1,300 AdaptMMBench samples, with auxiliary-line tasks evaluated separately. * indicates that the model supports enhancement operations. ``w/o enh.'' denotes results without enhancement-based data transformations (e.g., rotation and contrast).}
  \label{tab:accuracy_results}
  \begin{adjustbox}{width=0.95\textwidth}
    \begin{tabular}{ll cc cc cc cc cc | cc}
      \toprule
      \multirow{2}{*}{\textbf{Model}} & \multirow{2}{*}{\textbf{Mode}} & \multicolumn{2}{c}{\textbf{Real-world}} & \multicolumn{2}{c}{\textbf{OCR}}   & \multicolumn{2}{c}{\textbf{GUI}}   & \multicolumn{2}{c}{\textbf{Knowledge}} & \multicolumn{2}{c|}{\textbf{Math (w/o aux)}} & \multicolumn{2}{c}{\textbf{Overall Accuracy}}                                                                                                                                                                                                                               \\
      \cmidrule(lr){3-4} \cmidrule(lr){5-6} \cmidrule(lr){7-8} \cmidrule(lr){9-10} \cmidrule(lr){11-12} \cmidrule(l){13-14}
                                      &                                & w/o enh.                                & All                                & w/o enh.                           & All                                    & w/o enh.                           & All                                           & w/o enh.                           & All                                & w/o enh.                           & All                                & w/o enh.                           & All                                \\
                                        
      \midrule
      \rowcolor{headergray} \multicolumn{14}{c}{\textit{Open-Source Models}} \\
      \midrule
      \multirow{3}{*}{PixelReasoner}
                                      & \color{darkergray}Text         & \color{darkergray}42.08                 & \color{darkergray}38.00            & \color{darkergray}58.75            & \color{darkergray}58.33                & \color{darkergray}48.33            & \color{darkergray}48.00                       & \color{darkergray}56.88            & \color{darkergray}55.50            & \color{darkergray}46.25            & \color{darkergray}43.00            & \color{darkergray}50.29            & \color{darkergray}48.46            \\
                                      & Adaptive                       & 53.75                                   & 51.33                              & 61.25                              & 62.33                                  & 61.67                              & 53.00                                         & 58.13                              & 59.00                              & 51.88                              & 50.00                              & 55.19                              & 55.23                              \\
                                      & \color{darkergray}Oracle        & \color{darkergray}70.83                 & \color{darkergray}67.67            & \color{darkergray}72.92            & \color{darkergray}75.00                & \color{darkergray}66.67            & \color{darkergray}58.67                       & \color{darkergray}65.62            & \color{darkergray}67.50            & \color{darkergray}62.50            & \color{darkergray}64.00            & \color{darkergray}65.96            & \color{darkergray}66.69            \\
      \midrule
      \multirow{3}{*}{Deepeyes}
                                      & \color{darkergray}Text         & \color{darkergray}50.00                 & \color{darkergray}48.33            & \color{darkergray}50.42            & \color{darkergray}51.33                & \color{darkergray}52.50            & \color{darkergray}53.33                       & \color{darkergray}50.62            & \color{darkergray}48.50            & \color{darkergray}43.12            & \color{darkergray}41.50            & \color{darkergray}49.71            & \color{darkergray}49.15            \\
                                      & Adaptive                       & 54.17                                   & 53.00                              & 57.08                              & 57.67                                  & 51.25                              & 52.33                                         & 53.12                              & 50.00                              & 55.00                              & 50.50                              & 54.13                              & 53.08                              \\
                                      & \color{darkergray}Oracle        & \color{darkergray}67.08                 & \color{darkergray}66.33            & \color{darkergray}62.92            & \color{darkergray}66.67                & \color{darkergray}61.67            & \color{darkergray}64.00                       & \color{darkergray}66.88            & \color{darkergray}69.00            & \color{darkergray}61.88            & \color{darkergray}62.00            & \color{darkergray}64.04            & \color{darkergray}65.69            \\
      \midrule
      \multirow{3}{*}{Thyme$^*$}
                                      & \color{darkergray}Text         & \color{darkergray}55.00                 & \color{darkergray}51.67            & \color{darkergray}58.33            & \color{darkergray}57.67                & \color{darkergray}50.00            & \color{darkergray}49.67                       & \color{darkergray}55.00            & \color{darkergray}51.00            & \color{darkergray}51.88            & \color{darkergray}48.00            & \color{darkergray}54.13            & \color{darkergray}51.92            \\
                                      & Adaptive                       & 60.83                                   & 58.00                              & 61.67                              & 60.00                                  & 55.83                              & 53.67                                         & 58.75                              & 51.00                              & 51.88                              & 50.00                              & 58.17                              & 55.15                              \\
                                      & \color{darkergray}Oracle        & \color{darkergray}75.83                 & \color{darkergray}73.33            & \color{darkergray}66.67            & \color{darkergray}69.00                & \color{darkergray}63.75            & \color{darkergray}64.67                       & \color{darkergray}64.38            & \color{darkergray}67.50            & \color{darkergray}63.12            & \color{darkergray}63.00            & \color{darkergray}67.21            & \color{darkergray}67.85            \\
      \midrule
      \multirow{3}{*}{PyVision$^*$}
                                      & \color{darkergray}Text         & \color{darkergray}40.00                 & \color{darkergray}38.00            & \color{darkergray}62.08            & \color{darkergray}60.33                & \color{darkergray}75.00            & \color{darkergray}70.00                       & \color{darkergray}35.62            & \color{darkergray}34.50            & \color{darkergray}35.00            & \color{darkergray}31.00            & \color{darkergray}51.73            & \color{darkergray}48.92            \\
                                      & Adaptive                       & 50.83                                   & 47.33                              & 72.50                              & 70.67                                  & 77.92                              & 73.33                                         & 58.75                              & 52.50                              & 35.00                              & 31.00                              & 60.87                              & 57.00                              \\
                                      & \color{darkergray}Oracle        & \color{darkergray}94.58                 & \color{darkergray}84.67            & \color{darkergray}79.58            & \color{darkergray}80.67                & \color{darkergray}90.00            & \color{darkergray}86.00                       & \color{darkergray}60.00            & \color{darkergray}62.50            & \color{darkergray}58.13            & \color{darkergray}53.50            & \color{darkergray}79.13            & \color{darkergray}75.85            \\
      \midrule
      \multirow{3}{*}{Deepeyes v2$^*$}
                                      & \color{darkergray}Text         & \color{darkergray}58.75                 & \color{darkergray}55.00            & \color{darkergray}58.33            & \color{darkergray}58.33                & \color{darkergray}54.58            & \color{darkergray}54.67                       & \color{darkergray}48.12            & \color{darkergray}47.50            & \color{darkergray}41.88            & \color{darkergray}39.00            & \color{darkergray}53.46            & \color{darkergray}52.08            \\
                                      & Adaptive                       & 61.25                                   & 56.33                              & 59.58                              & 57.67                                  & 57.08                              & 55.67                                         & 59.38                              & 53.50                              & 50.00                              & 49.00                              & 57.88                              & 54.92                              \\
                                      & \color{darkergray}Oracle        & \color{darkergray}75.42                 & \color{darkergray}74.33            & \color{darkergray}70.83            & \color{darkergray}74.33                & \color{darkergray}68.33            & \color{darkergray}69.33                       & \color{darkergray}65.00            & \color{darkergray}65.50            & \color{darkergray}63.12            & \color{darkergray}64.00            & \color{darkergray}69.23            & \color{darkergray}70.23            \\
      \midrule
      \multirow{3}{*}{AdaptVision}
                                      & \color{darkergray}Text         & \color{darkergray}45.83                 & \color{darkergray}43.00            & \color{darkergray}62.92            & \color{darkergray}61.67                & \color{darkergray}48.75            & \color{darkergray}47.67                       & \color{darkergray}54.37            & \color{darkergray}54.00            & \color{darkergray}45.00            & \color{darkergray}44.50            & \color{darkergray}51.63            & \color{darkergray}50.31            \\
                                      & Adaptive                       & 49.17                                   & 46.33                              & 64.17                              & 64.00                                  & 52.92                              & 54.33                                         & 60.62                              & 54.50                              & 49.38                              & 45.00                              & 55.29                              & 53.31                              \\
                                      & \color{darkergray}Oracle        & \color{darkergray}74.17                 & \color{darkergray}70.67            & \color{darkergray}71.25            & \color{darkergray}76.00                & \color{darkergray}62.92            & \color{darkergray}65.67                       & \color{darkergray}71.88            & \color{darkergray}73.00            & \color{darkergray}68.75            & \color{darkergray}69.00            & \color{darkergray}69.71            & \color{darkergray}70.85            \\

      \midrule
      \multirow{3}{*}{\shortstack[l]{Qwen3-vl                                                                                                                                                                                                                                                                                                                                                                                                                                                                                                          \\-8B-Instruct}}
                                      & \color{darkergray}Text         & \color{darkergray}56.25                 & \color{darkergray}50.00            & \color{darkergray}64.17            & \color{darkergray}62.33                & \color{darkergray}57.50            & \color{darkergray}54.00                       & \color{darkergray}72.50            & \color{darkergray}66.00            & \color{darkergray}55.62            & \color{darkergray}50.50            & \color{darkergray}60.77            & \color{darkergray}56.31            \\
                                      & Adaptive                       & 57.50                                   & 52.33                              & 68.33                              & 65.67                                  & 65.83                              & 59.67                                         & 78.75                              & 71.50                              & 69.38                              & 62.00                              & 67.02                              & 61.54                              \\
                                      & \color{darkergray}Oracle        & \color{darkergray}83.75                 & \color{darkergray}78.00            & \color{darkergray}79.17            & \color{darkergray}80.67                & \color{darkergray}68.75            & \color{darkergray}67.00                       & \color{darkergray}80.62            & \color{darkergray}81.00            & \color{darkergray}80.00            & \color{darkergray}80.00            & \color{darkergray}78.17            & \color{darkergray}76.85            \\
      \midrule
      \multirow{3}{*}{\shortstack[l]{Qwen3-vl                                                                                                                                                                                                                                                                                                                                                                                                                                                                                                          \\-32B-Instruct}}
                                      & \color{darkergray}Text         & \color{darkergray}55.83                 & \color{darkergray}51.00            & \color{darkergray}82.92            & \color{darkergray}78.00                & \color{darkergray}76.67            & \color{darkergray}71.33                       & \color{darkergray}83.75            & \color{darkergray}77.00            & \color{darkergray}76.25            & \color{darkergray}68.00            & \color{darkergray}74.33            & \color{darkergray}68.54            \\
                                      & Adaptive                       & 63.33                                   & 57.33                              & 85.42                              & 82.67                                  & 77.00                              & 70.00                                         & 84.38                              & 79.00                              & 81.88                              & 73.50                              & 77.79                              & 71.92                              \\
                                      & \color{darkergray}Oracle        & \color{darkergray}87.08                 & \color{darkergray}81.67            & \color{darkergray}92.92            & \color{darkergray}92.00                & \color{darkergray}81.67            & \color{darkergray}75.67                       & \color{darkergray}96.25            & \color{darkergray}95.00            & \color{darkergray}90.00            & \color{darkergray}87.50            & \color{darkergray}89.04            & \color{darkergray}85.62            \\
      \midrule
      \multirow{3}{*}{\shortstack[l]{Qwen3-vl                                                                                                                                                                                                                                                                                                                                                                                                                                                                                                          \\-235B-Instruct}}
                                      & \color{darkergray}Text         & \color{darkergray}59.58                 & \color{darkergray}52.33            & \color{darkergray}81.25            & \color{darkergray}78.00                & \color{darkergray}84.58            & \color{darkergray}76.67                       & \color{darkergray}83.12            & \color{darkergray}78.50            & \color{darkergray}83.12            & \color{darkergray}73.00            & \color{darkergray}77.60            & \color{darkergray}71.08            \\
                                      & Adaptive                       & 64.17                                   & 56.33                              & 82.08                              & 80.00                                  & 87.08                              & 80.00                                         & \cellcolor{softblue}\textbf{93.75} & \cellcolor{softgreen}86.50         & \cellcolor{softgreen}85.62         & \cellcolor{softgreen}76.50         & 81.44                              & 75.00                              \\
                                      & \color{darkergray}Oracle        & \color{darkergray}87.92                 & \color{darkergray}80.67            & \color{darkergray}90.83            & \color{darkergray}91.33                & \color{darkergray}97.08            & \color{darkergray}91.33                       & \color{darkergray}96.25            & \color{darkergray}96.50            & \color{darkergray}96.88            & \color{darkergray}94.00            & \color{darkergray}93.37            & \color{darkergray}90.08            \\
      \midrule
      \rowcolor{headergray} \multicolumn{14}{c}{\textit{Closed-Source Moddels}}                                                                                                                                                                                                                                                                                                                                                                                                                                                                        \\
      \midrule
      \multirow{3}{*}{GPT-5$^*$}
                                      & \color{darkergray}Text         & \color{darkergray}46.67                 & \color{darkergray}45.67            & \color{darkergray}77.08            & \color{darkergray}73.67                & \color{darkergray}79.17            & \color{darkergray}74.33                       & \color{darkergray}60.00            & \color{darkergray}54.50            & \color{darkergray}44.38            & \color{darkergray}39.00            & \color{darkergray}62.88            & \color{darkergray}59.08            \\
                                      & Adaptive                       & \cellcolor{softgreen}70.83              & \cellcolor{softgreen}64.67         & \cellcolor{softgreen}89.17         & \cellcolor{softgreen}86.33             & \cellcolor{softgreen}88.75         & \cellcolor{softgreen}85.33                    & \cellcolor{softgreen}92.50         & 86.00                              & 76.88                              & 71.00                              & \cellcolor{softgreen}83.46         & \cellcolor{softgreen}78.69         \\
                                      & \color{darkergray}Oracle        & \color{darkergray}97.92                 & \color{darkergray}88.00            & \color{darkergray}90.42            & \color{darkergray}90.00                & \color{darkergray}91.67            & \color{darkergray}88.00                       & \color{darkergray}96.88            & \color{darkergray}94.50            & \color{darkergray}90.62            & \color{darkergray}83.00            & \color{darkergray}93.46            & \color{darkergray}88.69            \\
      \midrule
      \multirow{3}{*}{Gemini-3-Pro$^*$}
                                      & \color{darkergray}Text         & \color{darkergray}59.17                 & \color{darkergray}53.33            & \color{darkergray}87.92            & \color{darkergray}87.33                & \color{darkergray}90.83            & \color{darkergray}86.67                       & \color{darkergray}85.00            & \color{darkergray}83.00            & \color{darkergray}81.88            & \color{darkergray}75.50            & \color{darkergray}80.58            & \color{darkergray}76.85            \\
                                      & Adaptive                       & \cellcolor{softblue}\textbf{80.42}      & \cellcolor{softblue}\textbf{74.00} & \cellcolor{softblue}\textbf{89.58} & \cellcolor{softblue}\textbf{89.67}     & \cellcolor{softblue}\textbf{92.08} & \cellcolor{softblue}\textbf{90.00}            & \cellcolor{softgreen}92.50         & \cellcolor{softblue}\textbf{93.50} & \cellcolor{softblue}\textbf{93.12} & \cellcolor{softblue}\textbf{87.00} & \cellcolor{softblue}\textbf{89.04} & \cellcolor{softblue}\textbf{86.31} \\
                                      & \color{darkergray}Oracle        & \color{darkergray}87.50                 & \color{darkergray}80.33            & \color{darkergray}92.59            & \color{darkergray}93.00                & \color{darkergray}94.17            & \color{darkergray}92.67                       & \color{darkergray}92.50            & \color{darkergray}94.00            & \color{darkergray}93.75            & \color{darkergray}91.00            & \color{darkergray}91.92            & \color{darkergray}89.85            \\
      \bottomrule
    \end{tabular}
  \end{adjustbox}
\end{table*}

\subsection{Experiment Setting}
We evaluate a set of VLMs to establish baselines for \bench. For closed-source models, we select GPT-5~\cite{singh2025openai} and Gemini3~\cite{gemini3}. For open-source models, we include the Qwen3-VL family~\cite{Qwen3-VL} at multiple scales (8B, 32B, and 235B). In addition, we evaluate several specialized adaptive reasoning models, including DeepEyes~\cite{zheng2025deepeyes, hong2025deepeyesv2}, PixelReasoner~\cite{wang2025pixel}, Thyme~\cite{zhang2025thyme}, PyVision~\cite{zhao2025pyvision}, and AdaptVision~\cite{lin2025adaptvision}. 
For all evaluated models, we follow the implementation details provided in their official codebases. For evaluations under different reasoning modes, we apply a unified and minimal modification to the prompts, as detailed in the Appendix~\ref{sec:reasoning_mode_prompt}.

\subsection{Adaptive Reasoning Mode Selection Capability}
A closer analysis of mode selection capability reveals clear differences across models. As shown in Table~\ref{tab:adaptive_results} and Table~\ref{tab:accuracy_results}, mode selection capability does not exhibit a strong correlation with final task accuracy. For example, AdaptVision achieves a relatively modest accuracy, yet demonstrates strong mode selection behavior with an MCC of 0.17, outperforming all other models trained on Qwen2.5-VL-7B backbones. In contrast, GPT-5 attains the highest MCC of 0.41, demonstrating good mode selection capability.

\noindent\textit{\textbf{Model scaling improves mode selection.}}
Table~\ref{tab:adaptive_results} demonstrates a clear scaling trend within the Qwen3-VL family, where larger models exhibit more reliable mode selection. This pattern suggests that increased model capacity contributes to improved calibration when determining whether tool-based reasoning is necessary. Similarly, large-scale closed-source models outperform open-source models.

\noindent\textit{\textbf{Imbalanced mode selection behavior is observed in some models.}}
Several specialized adaptive models exhibit imbalanced mode selection behavior, either invoking tools excessively or rarely. For example, Deepeyes v2 invokes tools in all but one of the 1,300 samples in \bench, whereas Thyme triggers tool usage in only about 3\% of cases. Such imbalanced patterns are associated with lower mode selection performance, despite competitive accuracy.

\subsection{Quality and Efficiency of the Reasoning Process}
Since intermediate reasoning steps of closed-source models (e.g., GPT-5 and Gemini-3-Pro) are not accessible, we restrict process-level analysis to open-source models. Table~\ref{tab:process_results} evaluates key step coverage, tool effectiveness, and efficiency. Consistent with Table~\ref{tab:accuracy_results}, key step coverage shows a similar ranking, with Qwen3-VL-235B among the top models. Larger models also demonstrate stronger tool effectiveness, better aligning tool usage with reasoning intent.

\noindent\textit{\textbf{Tool effectiveness varies with models.}}
Qwen3 family shows strong performance, while some smaller models are less effective. This may stem from repeated or unnecessary tool calls, as well as code-based tool invocation in Deepeyes v2, Thyme, and PyVision, which introduces more complexity than the function-call interface used by Qwen models.

\noindent\textit{\textbf{Token usage is not positively correlated with steps or tool calls.}}
Considering efficiency, token usage varies across models and does not correspond to the number of reasoning steps or tool calls. For example, Thyme uses the fewest steps and tool invocations, yet consumes more tokens than PyVision, which has the most steps. This shows that fewer steps or tool calls do not necessarily reduce token cost.

\subsection{Accuracy across Reasoning Modes}
We analyze model performance across different reasoning modes, including text-only, adaptive, and oracle tool reasoning. The oracle tool reflects upper-bound performance. As shown in Table~\ref{tab:accuracy_results}, adaptive reasoning consistently improves accuracy over text-only baselines for all evaluated models.

\noindent\textit{\textbf{Significant performance gap between adaptive and oracle reasoning.}}
Although adaptive reasoning yields clear gains, oracle tool reasoning reveals substantial remaining headroom. For example, GPT-5 improves from 78.69\% under adaptive reasoning to 88.69\% in the oracle setting, with similar trends observed in open-source models. 
These results indicate that current performance is mainly limited by imperfect tool invocation rather than reasoning capability. Moreover, the high oracle-visual accuracy of 90.08\% indicates the reliability and accuracy of our visual annotations.

\noindent\textit{\textbf{Generation-Based Tools Are Beneficial for Certain Tasks.}}
We conduct an exploratory analysis on self-generated auxiliary-line tasks as shown in Table~\ref{tab:math_aux}. As current open-source models cannot generate visual representations, adaptive reasoning shows limited or negative gains over text-only reasoning, while oracle-visual inputs bring substantial improvements. This highlights the importance of visual generation for future adaptive reasoning models.

\begin{table}[ht!]
  \centering
  \definecolor{softblue}{rgb}{0.93, 0.96, 1.0}
  \definecolor{softgreen}{rgb}{0.94, 0.98, 0.94}
  \definecolor{reasoningyellow}{rgb}{1.0, 0.98, 0.9}
  \definecolor{opensourcecyan}{rgb}{0.9, 0.98, 1.0}
  \definecolor{closedpurple}{rgb}{0.96, 0.94, 1.0}
  \definecolor{headergray}{rgb}{0.95, 0.95, 0.95}
  \definecolor{darkergray}{rgb}{0.35, 0.35, 0.35}

  \caption{Experimental results on geometric auxiliary-line problems across different reasoning modes.}
  \label{tab:math_aux}
  \begin{adjustbox}{width=0.36\textwidth}
    \begin{tabular}{l ccc}
      \toprule
      \textbf{Model} & \textbf{Text Acc} & \textbf{Adaptive Acc} & \textbf{Oracle Acc} \\
      \midrule
      \multicolumn{4}{c}{\textit{Open-Source VLMs}}                                  \\
      \midrule
      Thyme          & 21.67             & 21.67                 & 24.17             \\
      PyVision       & 15.83             & 29.17     & 32.50             \\
      Deepeyes v2   & 19.17             & 19.17                 & 25.83             \\
      Qwen3-vl-8B    & 50.00             & 46.67                 & 62.50             \\
      Qwen3-vl-32B   & 63.33    & 58.33                 & 79.17 \\
      Qwen3-vl-235B  & 62.50 & 68.33        & 84.17    \\
      \midrule
      \multicolumn{4}{c}{\textit{Closed-Source Models}}                              \\
      \midrule
      Gemini-3-Pro         & \cellcolor{softblue}\textbf{85.00}    & \cellcolor{softgreen}{78.33}     & \cellcolor{softblue}\textbf{94.17}    \\
      GPT-5          & \cellcolor{softgreen}{75.00} & \cellcolor{softblue}\textbf{86.67}        & \cellcolor{softgreen}{89.17} \\
      \bottomrule
    \end{tabular}
  \end{adjustbox}
\end{table}

\subsection{Error Analysis}
\label{sec:error_analysys}
\begin{figure}[ht!]
  \centering
  \includegraphics[width=0.7\linewidth]{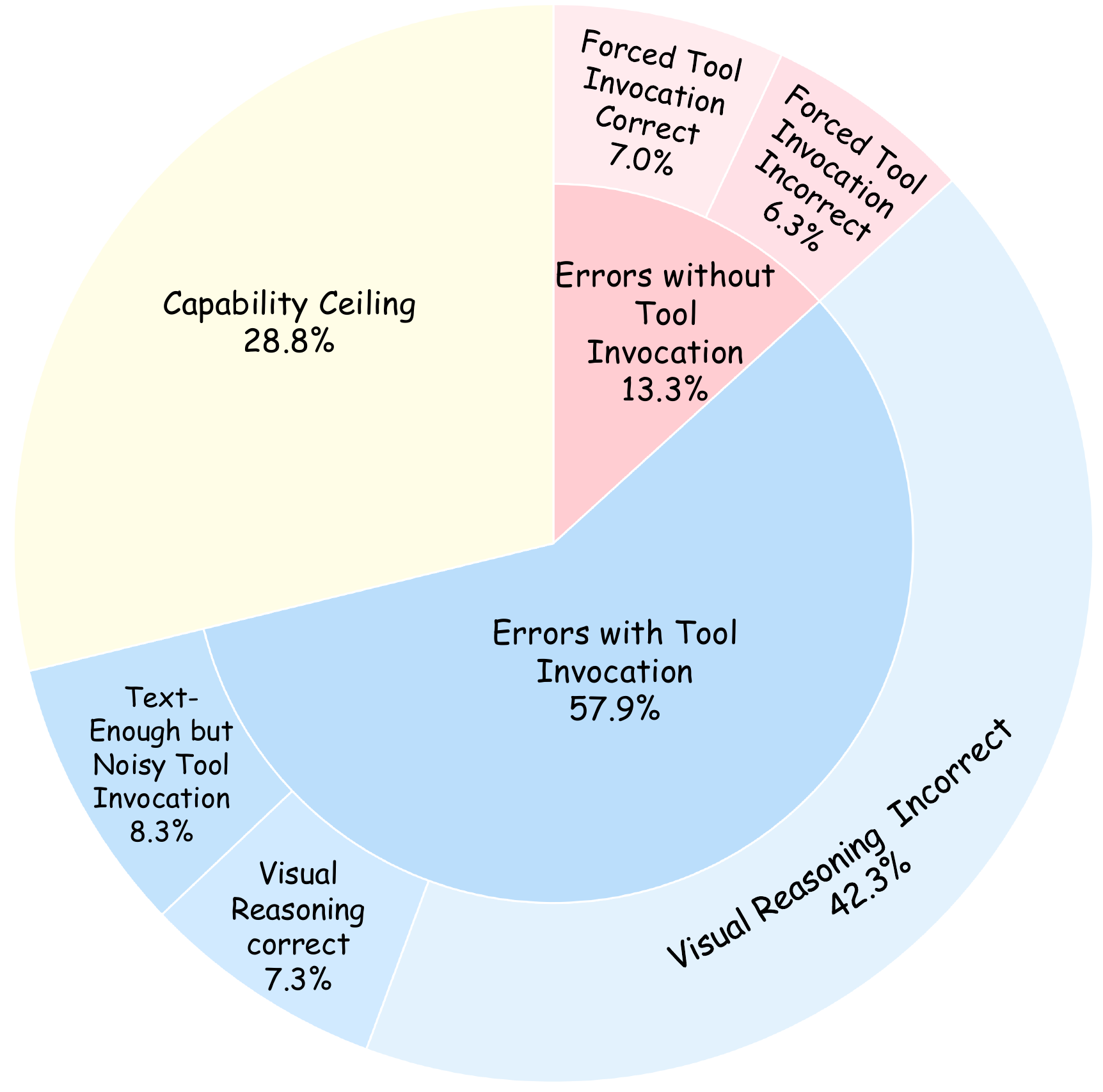}
  \caption{Error Analysis on GPT-5.}
  \label{fig:failure_gpt}
\end{figure}

In this section, we analyze the causes of incorrect predictions made by GPT-5 under the adaptive mode to understand the gap between adaptive reasoning and oracle-visual mode. As shown in Fig.~\ref{fig:failure_gpt}, most errors are related to tool usage. Specifically, 42.3\% of the errors stem from visual reasoning failures, such as zoom-in into incorrect regions or applying wrong image transformations. Another 7.3\% of errors occur even when visual reasoning is correct. Since these samples are solvable in the oracle-visual mode, this suggests that intermediate images in multi-step reasoning may introduce visual noise affecting the final prediction. In addition, 8.3\% of errors are caused by incorrect mode selection, where text reasoning is sufficient but the model unnecessarily invokes tools, leading to degraded performance. For cases without tool usage, forcing tool invocation corrects 7.0\% of the errors, while 6.3\% remain incorrect. The remaining 28.8\% of errors exceed the capability of the GPT-5 model.

\section{Conclusion}
In this paper, we present \bench, a benchmark for evaluating adaptive multimodal reasoning in VLMs. \bench\ covers diverse domains and reasoning scenarios, and enables model-dependent identification of tool-redundant and tool-required cases by comparing performance across reasoning modes. We further propose a set of metrics that assess mode selection quality, reasoning process quality, and efficiency. Through systematic evaluation of state-of-the-art models, we observe that high accuracy does not necessarily imply strong reasoning mode selection capability. The substantial performance gap between adaptive and oracle-visual reasoning further suggests that performance is often limited by suboptimal tool invocation. This highlights adaptive tool selection as a key challenge for future multimodal reasoning models.


% \section*{Impact Statement}

% Authors are \textbf{required} to include a statement of the potential broader
% impact of their work, including its ethical aspects and future societal
% consequences. This statement should be in an unnumbered section at the end of
% the paper (co-located with Acknowledgements -- the two may appear in either
% order, but both must be before References), and does not count toward the paper
% page limit. In many cases, where the ethical impacts and expected societal
% implications are those that are well established when advancing the field of
% Machine Learning, substantial discussion is not required, and a simple
% statement such as the following will suffice:

% ``This paper presents work whose goal is to advance the field of Machine
% Learning. There are many potential societal consequences of our work, none
% which we feel must be specifically highlighted here.''

% The above statement can be used verbatim in such cases, but we encourage
% authors to think about whether there is content which does warrant further
% discussion, as this statement will be apparent if the paper is later flagged
% for ethics review.

\section*{Impact Statement}
This paper presents work whose goal is to advance the field of Machine Learning. There are many potential societal consequences of our work, none which we feel must be specifically highlighted here.

\bibliography{ref}
\bibliographystyle{icml2026}

%%%%%%%%%%%%%%%%%%%%%%%%%%%%%%%%%%%%%%%%%%%%%%%%%%%%%%%%%%%%%%%%%%%%%%%%%%%%%%%
%%%%%%%%%%%%%%%%%%%%%%%%%%%%%%%%%%%%%%%%%%%%%%%%%%%%%%%%%%%%%%%%%%%%%%%%%%%%%%%
% APPENDIX
%%%%%%%%%%%%%%%%%%%%%%%%%%%%%%%%%%%%%%%%%%%%%%%%%%%%%%%%%%%%%%%%%%%%%%%%%%%%%%%
%%%%%%%%%%%%%%%%%%%%%%%%%%%%%%%%%%%%%%%%%%%%%%%%%%%%%%%%%%%%%%%%%%%%%%%%%%%%%%%
% \newpage
% \appendix
% \onecolumn

% You can have as much text here as you want. The main body must be at most $8$
% pages long. For the final version, one more page can be added. If you want, you
% can use an appendix like this one.

% The $\mathtt{\backslash onecolumn}$ command above can be kept in place if you
% prefer a one-column appendix, or can be removed if you prefer a two-column
% appendix.  Apart from this possible change, the style (font size, spacing,
% margins, page numbering, etc.) should be kept the same as the main body.

\newpage
\appendix
\onecolumn

% You can have as much text here as you want. The main body must be at most $8$
% pages long. For the final version, one more page can be added. If you want, you
% can use an appendix like this one.

% The $\mathtt{\backslash onecolumn}$ command above can be kept in place if you
% prefer a one-column appendix, or can be removed if you prefer a two-column
% appendix.  Apart from this possible change, the style (font size, spacing,
% margins, page numbering, etc.) should be kept the same as the main body.

\section{More Data Details}
\label{sec:Data_Source_Distribution}

\subsection{Data Source Distribution}
\begin{figure*}[ht!]
    \centering
    \includegraphics[width=0.9\linewidth]{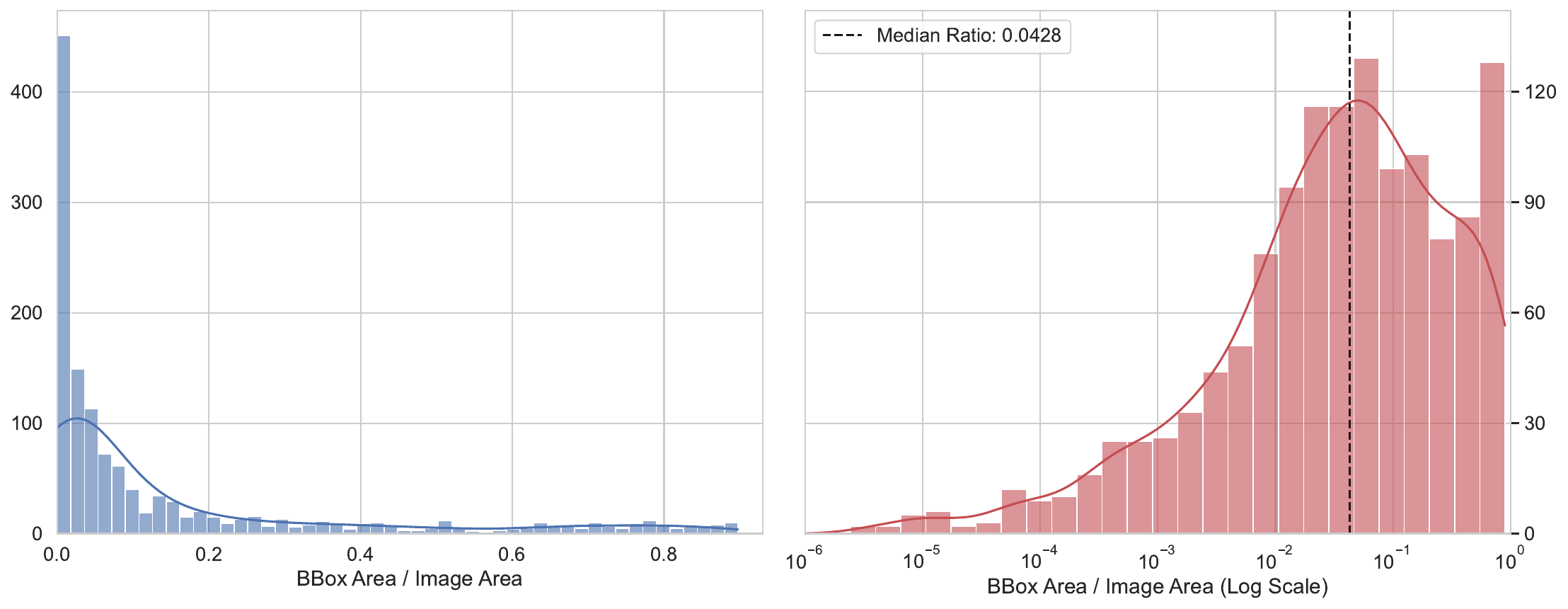}
     \caption{Statistics of bounding box sizes in \bench.}
    \label{fig:bbox_stats}
\end{figure*}

\noindent\textbf{Real-World VQA.} We target high-resolution natural scenes by leveraging VisualProbe~\cite{lai2025mini} for small-object search and a custom SA-1B~\cite{kirillov2023segment} subset for large-scale object reasoning. Queries are explicitly designed to evaluate attributes, spatial, counting, physical state and text-recognition across distinct scales. Moreover, statistics of bounding box sizes are presented in Fig. \ref{fig:bbox_stats}.

\noindent\textbf{Text-Rich VQA.} This domain covers diverse charts, tables, and documents. We aggregate standard samples from ChartQA~\cite{masry2022chartqa} and DocVQA~\cite{mathew2021docvqa} with high-resolution challenges from ChartQA-Pro~\cite{masry2025chartqapro}, MM-RealWorld~\cite{zhang2025mme}, and Insight-o3~\cite{li2025insight} to demand precise visual inspection and deep reasoning.

\noindent\textbf{Math VQA.} To assess mathematical reasoning in visual contexts, we consolidate high-quality samples from a spectrum of established benchmarks including MathVista~\cite{lu2023mathvista}, MathVerse~\cite{zhang2024mathverse}, We-Math~\cite{qiao2025we}, LogicVista~\cite{xiao2024logicvista}, Visulogic~\cite{xu2025visulogic}, AuxSolidMath, and VTBench~\cite{lin2025vtbench}.

\noindent\textbf{GUI VQA.} We construct a cross-platform suite covering iOS, Android, Web, macOS, Windows, and Linux. This is achieved by integrating generic datasets like GUI-Knowledge-Bench~\cite{shi2025gui} and MMBench-GUI~\cite{liu2024mmbench} with domain-specific samples from WebWalker~\cite{wu2025webwalker}.
% and VisualWebBench.

\noindent\textbf{Knowledge VQA.} This category is sourced from disciplinary benchmarks across Physics, Chemistry, and Biology. Specifically, we incorporate expert-level samples from \textbf{MMMU}~\cite{yue2024mmmu} and \textbf{SciVerse}~\cite{guo2025sciverse} to evaluate the models' ability to integrate specialized domain knowledge with visual reasoning.

\begin{figure*}[ht!]
  \centering
  \includegraphics[width=0.8\linewidth]{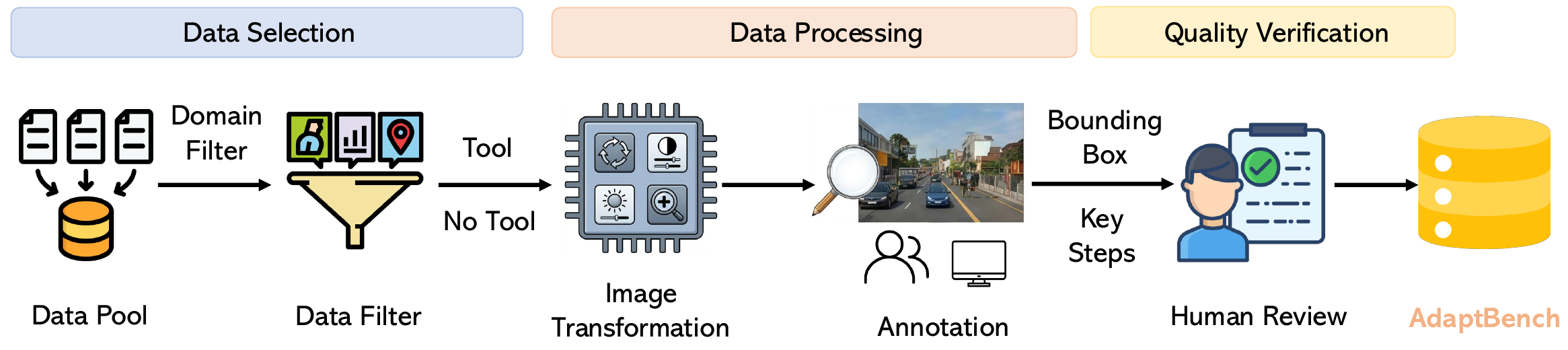}
  \caption{Overview of the data curation and annotation process.}
  \label{fig:data_pipeline}
\end{figure*}

\subsection{Data Construction Pipeline}

The construction workflow of AdaptBench is depicted in Figure~\ref{fig:data_pipeline}. Initially, raw data is partitioned based on reasoning complexity, specifically separating tasks that necessitate external tool intervention from those amenable to text-only inference. To further challenge model adaptability, we augment the visual inputs with diverse transformations, thereby mandating fine-grained perception. Finally, we implement a multi-stage verification pipeline involving expert annotation of transformation logic, key reasoning steps, and rigorous human review, ensuring a high-fidelity ground truth for the final benchmark.

% \newpage
\section{Transform Results}

Tab.~\ref{tab:transformed_acc} serves as a supplementary detailed analysis to the main experimental results presented in Tab.~\ref{tab:accuracy_results}. While Tab.~\ref{tab:accuracy_results} reports the model performance on the original dataset (denoted as ``w/o enh.'') and the aggregated dataset (``All''), it does not explicitly isolate the performance on the transformed data. To provide a comprehensive view of model robustness against data variations, Tab.~\ref{tab:transformed_acc} exclusively presents the accuracy results for the \textbf{Transformed} subset across all five domains. The highlighting scheme follows the same convention as the main table to facilitate direct comparison of the autonomous decision-making capabilities in the Adaptive mode.

\begin{table*}[ht!]
  \centering
  \definecolor{softblue}{rgb}{0.93, 0.96, 1.0}
  \definecolor{softgreen}{rgb}{0.94, 0.98, 0.94}
  \definecolor{headergray}{rgb}{0.95, 0.95, 0.95}
  \definecolor{darkergray}{rgb}{0.35, 0.35, 0.35}

  \caption{Main experimental results on the \textbf{Transform} subset of AdaptBench. We report accuracy (\%) across five specific domains and overall aggregates. Performance for \textbf{Text} and \textbf{Oracle} modes is displayed in \textcolor{darkergray}{deep gray} to prioritize adaptive results. * indicates that the model supports enhancement operations.}
  
  \label{tab:transformed_acc}
  
  \begin{adjustbox}{width=0.8\textwidth}
    \begin{tabular}{ll ccccc|c}
      \toprule
      \textbf{Model} & \textbf{Mode} & \textbf{Real-world} & \textbf{OCR} & \textbf{GUI} & \textbf{Knowledge} & \textbf{Math} & \textbf{Overall Accuracy} \\
      \midrule

      \rowcolor{headergray} \multicolumn{8}{c}{\textit{Open-Source Models}} \\
      \midrule
      
        \multirow{3}{*}{PixelReasoner}
            & \color{darkergray}Text     & \color{darkergray}46.67 & \color{darkergray}50.00 & \color{darkergray}30.00 & \color{darkergray}56.67 & \color{darkergray}21.67 & \color{darkergray}41.15 \\
            & Adaptive                   & 61.67 & 62.50 & 42.50 & 66.67 & 41.67 & 55.38 \\
            & \color{darkergray}Oracle     & \color{darkergray}66.67 & \color{darkergray}75.00 & \color{darkergray}70.00 & \color{darkergray}83.33 & \color{darkergray}55.00 & \color{darkergray}69.62 \\
        \midrule

        \multirow{3}{*}{Deepeyes}
            & \color{darkergray}Text     & \color{darkergray}56.67 & \color{darkergray}40.00 & \color{darkergray}35.00 & \color{darkergray}55.00 & \color{darkergray}41.67 & \color{darkergray}46.92 \\
            & Adaptive                   & 56.67 & 37.50 & 32.50 & 60.00 & 48.33 & 48.85 \\
            & \color{darkergray}Oracle     & \color{darkergray}73.33 & \color{darkergray}77.50 & \color{darkergray}65.00 & \color{darkergray}81.67 & \color{darkergray}63.33 & \color{darkergray}72.31 \\
        \midrule

        \multirow{3}{*}{Thyme$^*$}
            & \color{darkergray}Text     & \color{darkergray}48.33 & \color{darkergray}35.00 & \color{darkergray}32.50 & \color{darkergray}55.00 & \color{darkergray}38.33 & \color{darkergray}43.08 \\
            & Adaptive                   & 45.00 & 20.00 & 42.50 & 53.33 & 46.67 & 43.08 \\
            & \color{darkergray}Oracle     & \color{darkergray}68.33 & \color{darkergray}80.00 & \color{darkergray}62.50 & \color{darkergray}78.33 & \color{darkergray}63.33 & \color{darkergray}70.38 \\
        \midrule

        \multirow{3}{*}{PyVision$^*$}
            & \color{darkergray}Text     & \color{darkergray}50.00 & \color{darkergray}30.00 & \color{darkergray}15.00 & \color{darkergray}53.33 & \color{darkergray}30.00 & \color{darkergray}37.69 \\
            & Adaptive                   & 55.00 & 27.50 & 15.00 & 63.33 & 33.33 & 41.54 \\
            & \color{darkergray}Oracle     & \color{darkergray}70.00 & \color{darkergray}72.50 & \color{darkergray}35.00 & \color{darkergray}85.00 & \color{darkergray}45.00 & \color{darkergray}62.69 \\
        \midrule

        \multirow{3}{*}{Deepeyes v2$^*$}
            & \color{darkergray}Text     & \color{darkergray}55.00 & \color{darkergray}45.00 & \color{darkergray}27.50 & \color{darkergray}58.33 & \color{darkergray}40.00 & \color{darkergray}46.54 \\
            & Adaptive                   & 50.00 & 30.00 & 45.00 & 50.00 & 36.67 & 43.08 \\
            & \color{darkergray}Oracle     & \color{darkergray}73.33 & \color{darkergray}67.50 & \color{darkergray}67.50 & \color{darkergray}88.33 & \color{darkergray}70.00 & \color{darkergray}74.23 \\
        \midrule

        \multirow{3}{*}{AdaptVision}
            & \color{darkergray}Text     & \color{darkergray}43.33 & \color{darkergray}52.50 & \color{darkergray}42.50 & \color{darkergray}56.67 & \color{darkergray}31.67 & \color{darkergray}45.00 \\
            & Adaptive                   & 60.00 & 30.00 & 27.50 & 63.33 & 35.00 & 45.38 \\
            & \color{darkergray}Oracle     & \color{darkergray}76.67 & \color{darkergray}77.50 & \color{darkergray}70.00 & \color{darkergray}95.00 & \color{darkergray}56.67 & \color{darkergray}75.38 \\
        \midrule

      \multirow{3}{*}{\shortstack[l]{Qwen3-VL \\-8B-Instruct}}
        & \color{darkergray}Text     & \color{darkergray}40.00 & \color{darkergray}40.00 & \color{darkergray}30.00 & \color{darkergray}55.00 & \color{darkergray}25.00 & \color{darkergray}38.46 \\
        & Adaptive                   & 35.00 & 42.50 & 32.50 & 55.00 & 31.67 & 39.62 \\
        & \color{darkergray}Oracle     & \color{darkergray}60.00 & \color{darkergray}82.50 & \color{darkergray}80.00 & \color{darkergray}86.67 & \color{darkergray}55.00 & \color{darkergray}71.54 \\
      \midrule

      \multirow{3}{*}{\shortstack[l]{Qwen3-VL \\-32B-Instruct}}
        & \color{darkergray}Text     & \color{darkergray}50.00 & \color{darkergray}50.00 & \color{darkergray}35.00 & \color{darkergray}58.33 & \color{darkergray}31.67 & \color{darkergray}45.38 \\
            & Adaptive                   & 40.00 & 57.50 & 40.00 & 71.67 & 33.33 & 48.46 \\
            & \color{darkergray}Oracle     & \color{darkergray}51.67 & \color{darkergray}90.00 & \color{darkergray}77.50 & \color{darkergray}88.33 & \color{darkergray}60.00 & \color{darkergray}71.92 \\
      \midrule

      \multirow{3}{*}{\shortstack[l]{Qwen3-VL \\-235B-Instruct}}
        & \color{darkergray}Text     & \color{darkergray}45.00 & \color{darkergray}60.00 & \color{darkergray}32.50 & \color{darkergray}65.00 & \color{darkergray}23.33 & \color{darkergray}45.00 \\
            & Adaptive                   & 51.67 & 57.50 & 40.00 & 71.67 & 25.00 & 49.23 \\
            & \color{darkergray}Oracle     & \color{darkergray}68.33 & \color{darkergray}97.50 & \color{darkergray}82.50 & \color{darkergray}93.33 & \color{darkergray}51.67 & \color{darkergray}76.92 \\

      \midrule
      \rowcolor{headergray} \multicolumn{8}{c}{\textit{Closed-Source Models}} \\
      \midrule

      \multirow{3}{*}{GPT-5$^*$}
        & \color{darkergray}Text     & \color{darkergray}55.00 & \color{darkergray}32.50 & \color{darkergray}17.50 & \color{darkergray}60.00 & \color{darkergray}41.67 & \color{darkergray}43.85 \\
            & Adaptive                   & 71.67 & 60.00 & 47.50 & 75.00 & 40.00 & 59.62 \\
            & \color{darkergray}Oracle     & \color{darkergray}73.33 & \color{darkergray}85.00 & \color{darkergray}52.50 & \color{darkergray}88.33 & \color{darkergray}48.33 & \color{darkergray}69.62 \\
    \midrule

      \multirow{3}{*}{Gemini-3-Pro$^*$}
        & \color{darkergray}Text     & \color{darkergray}70.00 & \color{darkergray}75.00 & \color{darkergray}50.00 & \color{darkergray}85.00 & \color{darkergray}30.00 & \color{darkergray}61.92 \\
        & Adaptive                   & 81.67 & 97.50 & 62.50 & 90.00 & 48.33 & 75.38 \\
        & \color{darkergray}Oracle     & \color{darkergray}86.67 & \color{darkergray}100.00 & \color{darkergray}80.00 & \color{darkergray}95.00 & \color{darkergray}51.67 & \color{darkergray}81.54 \\

      \bottomrule
    \end{tabular}
  \end{adjustbox}
\end{table*}

\newpage
\section{Category Results}

In this section, we provide a fine-grained analysis of model performance across all specific categories defined in our benchmark. To ensure legibility and accommodate the wide range of sub-domains, the detailed accuracy results are presented in two separate tables:

\begin{itemize}
    \item \textbf{Tab.~\ref{tab:category_part1}} reports the performance metrics for the \textbf{GUI} and \textbf{Realworld} domains.
    \item \textbf{Tab.~\ref{tab:category_part2}} covers the \textbf{Knowledge}, \textbf{Math}, and \textbf{OCR} domains.
\end{itemize}

In both tables, the first row (labeled as \textit{N}) denotes the number of test samples available for each corresponding category. All accuracy values are reported in decimal format.

\begin{table*}[htbp]
\centering
\footnotesize 
\setlength{\tabcolsep}{5pt}
\caption{Detailed Accuracy (Part 1/2): GUI and Realworld domains. * indicates that the model supports enhancement operations.}
\label{tab:category_part1}

  \begin{tabular}{ll cccccc ccccc}
    \toprule
     & & \multicolumn{6}{c}{\textbf{GUI}} 
     & \multicolumn{5}{c}{\textbf{Realworld}} \\
    \cmidrule(lr){3-8} \cmidrule(lr){9-13}
    
    \textbf{Model} & \textbf{Mode} & 
    And. & Lin. & Mac. & Web & Win. & iOS & 
    Attr. & Count & Integ. & Phys. & Spat. \\
    \midrule
    
    \textit{N} & (Count) & 83 & 22 & 39 & 75 & 51 & 30 & 118 & 16 & 121 & 10 & 35 \\
    \midrule
    
    \multirow{3}{*}{PixelReasoner} 
     & Text & 0.51 & 0.45 & 0.31 & 0.52 & 0.53 & 0.47 & 0.45 & 0.12 & 0.35 & 0.50 & 0.34 \\
     & Adaptive & 0.51 & 0.50 & 0.41 & 0.61 & 0.57 & 0.50 & 0.55 & 0.50 & 0.49 & 0.70 & 0.43 \\
     & Oracle & 0.59 & 0.68 & 0.36 & 0.63 & 0.67 & 0.57 & 0.68 & 0.75 & 0.69 & 0.50 & 0.63 \\
    \midrule
    
    \multirow{3}{*}{Deepeyes}
     & Text & 0.59 & 0.45 & 0.36 & 0.63 & 0.45 & 0.57 & 0.55 & 0.38 & 0.44 & 0.30 & 0.51 \\
     & Adaptive & 0.48 & 0.36 & 0.44 & 0.68 & 0.49 & 0.53 & 0.58 & 0.44 & 0.50 & 0.30 & 0.57 \\
     & Oracle & 0.67 & 0.50 & 0.46 & 0.75 & 0.63 & 0.63 & 0.65 & 0.69 & 0.69 & 0.40 & 0.69 \\
    \midrule
    
    \multirow{3}{*}{Thyme$^*$}
     & Text & 0.53 & 0.45 & 0.36 & 0.60 & 0.41 & 0.50 & 0.57 & 0.62 & 0.51 & 0.40 & 0.34 \\
     & Adaptive & 0.54 & 0.55 & 0.41 & 0.64 & 0.45 & 0.57 & 0.64 & 0.75 & 0.54 & 0.40 & 0.51 \\
     & Oracle & 0.61 & 0.68 & 0.56 & 0.80 & 0.55 & 0.60 & 0.73 & 0.75 & 0.79 & 0.60 & 0.60 \\
    \midrule
    
    \multirow{3}{*}{PyVision$^*$}
     & Text & 0.81 & 0.64 & 0.82 & 0.47 & 0.76 & 0.77 & 0.51 & 0.44 & 0.24 & 0.40 & 0.40 \\
     & Adaptive & 0.83 & 0.59 & 0.82 & 0.51 & 0.80 & 0.90 & 0.58 & 0.44 & 0.36 & 0.50 & 0.51 \\
     & Oracle & 0.92 & 0.77 & 0.92 & 0.79 & 0.88 & 0.83 & 0.87 & 0.88 & 0.80 & 0.70 & 0.94 \\
    \midrule
    
    \multirow{3}{*}{Deepeyes v2$^*$}
     & Text & 0.46 & 0.45 & 0.46 & 0.68 & 0.59 & 0.57 & 0.60 & 0.50 & 0.52 & 0.70 & 0.46 \\
     & Adaptive & 0.49 & 0.45 & 0.44 & 0.71 & 0.53 & 0.63 & 0.58 & 0.69 & 0.55 & 0.50 & 0.49 \\
     & Oracle & 0.64 & 0.68 & 0.51 & 0.77 & 0.76 & 0.77 & 0.72 & 0.81 & 0.79 & 0.80 & 0.63 \\
    \midrule
    
    \multirow{3}{*}{AdaptVision}
     & Text & 0.48 & 0.32 & 0.28 & 0.59 & 0.49 & 0.53 & 0.48 & 0.25 & 0.40 & 0.60 & 0.40 \\
     & Adaptive & 0.51 & 0.36 & 0.41 & 0.68 & 0.55 & 0.60 & 0.54 & 0.25 & 0.40 & 0.60 & 0.46 \\
     & Oracle & 0.65 & 0.50 & 0.51 & 0.76 & 0.71 & 0.63 & 0.69 & 0.69 & 0.75 & 0.80 & 0.60 \\
    \midrule
    
    \multirow{3}{*}{\shortstack[l]{Qwen3-VL \\-8B-Instruct}}
     & Text & 0.52 & 0.45 & 0.49 & 0.64 & 0.51 & 0.53 & 0.55 & 0.44 & 0.47 & 0.70 & 0.40 \\
     & Adaptive & 0.61 & 0.64 & 0.46 & 0.63 & 0.57 & 0.67 & 0.58 & 0.50 & 0.44 & 0.70 & 0.57 \\
     & Oracle & 0.70 & 0.59 & 0.51 & 0.77 & 0.67 & 0.60 & 0.70 & 0.88 & 0.84 & 0.80 & 0.77 \\
    \midrule
    
    \multirow{3}{*}{\shortstack[l]{Qwen3-VL \\-32B-Instruct}}
     & Text & 0.81 & 0.77 & 0.56 & 0.75 & 0.57 & 0.77 & 0.56 & 0.56 & 0.50 & 0.50 & 0.34 \\
     & Adaptive & 0.76 & 0.77 & 0.51 & 0.68 & 0.67 & 0.83 & 0.64 & 0.50 & 0.54 & 0.80 & 0.46 \\
     & Oracle & 0.80 & 0.68 & 0.62 & 0.83 & 0.73 & 0.77 & 0.78 & 0.81 & 0.86 & 0.70 & 0.83 \\
    \midrule
    
    \multirow{3}{*}{\shortstack[l]{Qwen3-VL \\-235B-Instruct}}
     & Text & 0.80 & 0.77 & 0.79 & 0.65 & 0.88 & 0.73 & 0.61 & 0.50 & 0.50 & 0.40 & 0.34 \\
     & Adaptive & 0.83 & 0.77 & 0.87 & 0.76 & 0.78 & 0.77 & 0.60 & 0.50 & 0.57 & 0.50 & 0.46 \\
     & Oracle & 0.92 & 0.91 & 0.95 & 0.88 & 0.94 & 0.90 & 0.81 & 0.75 & 0.83 & 0.60 & 0.83 \\
    \midrule
    
    \multirow{3}{*}{GPT-5$^*$}
     & Text & 0.86 & 0.82 & 0.87 & 0.47 & 0.75 & 0.90 & 0.62 & 0.44 & 0.31 & 0.40 & 0.46 \\
     & Adaptive & 0.89 & 0.86 & 0.90 & 0.75 & 0.88 & 0.90 & 0.77 & 0.62 & 0.55 & 0.60 & 0.57 \\
     & Oracle & 0.96 & 0.95 & 0.82 & 0.84 & 0.80 & 0.90 & 0.92 & 0.88 & 0.82 & 0.80 & 0.97 \\
    \midrule
    
    \multirow{3}{*}{Gemini-3-Pro$^*$}
     & Text & 0.90 & 0.86 & 0.92 & 0.79 & 0.86 & 0.90 & 0.58 & 0.38 & 0.50 & 0.50 & 0.54 \\
     & Adaptive & 0.93 & 0.86 & 0.95 & 0.84 & 0.90 & 0.93 & 0.82 & 0.69 & 0.69 & 0.90 & 0.60 \\
     & Oracle & 0.94 & 0.91 & 0.97 & 0.91 & 0.90 & 0.93 & 0.83 & 0.75 & 0.78 & 0.70 & 0.86 \\
      \bottomrule
    \end{tabular}
\end{table*}

\begin{table*}[htbp]
\centering
\footnotesize
\setlength{\tabcolsep}{5pt}
\caption{Detailed Accuracy (Part 2/2): Knowledge, Math, and OCR domains. * indicates that the model supports enhancement operations.}
\label{tab:category_part2}

\begin{tabular}{ll cccc cccc cccc}
\toprule
 & & \multicolumn{4}{c}{\textbf{Knowledge}} 
 & \multicolumn{4}{c}{\textbf{Math}} 
 & \multicolumn{4}{c}{\textbf{OCR}} \\
\cmidrule(lr){3-6} \cmidrule(lr){7-10} \cmidrule(lr){11-14}

\textbf{Model} & \textbf{Mode} & 
Bio. & Chem. & Geo. & Phys. & 
Alg. & Geo. & Log. & Stat. & 
Chart & Diag. & Doc. & Tab. \\
\midrule

\textit{N} & (Count) & 57 & 58 & 10 & 75 & 64 & 75 & 13 & 48 & 171 & 40 & 55 & 34 \\
\midrule

\multirow{3}{*}{PixelReasoner} 
 & Text & 0.68 & 0.47 & 0.70 & 0.51 & 0.44 & 0.44 & 0.38 & 0.42 & 0.57 & 0.70 & 0.53 & 0.62 \\
 & Adaptive & 0.70 & 0.45 & 0.50 & 0.63 & 0.52 & 0.49 & 0.38 & 0.52 & 0.65 & 0.62 & 0.55 & 0.62 \\
 & Oracle & 0.82 & 0.62 & 0.60 & 0.61 & 0.69 & 0.69 & 0.46 & 0.54 & 0.72 & 0.88 & 0.73 & 0.79 \\
\midrule

\multirow{3}{*}{Deepeyes}
 & Text & 0.61 & 0.38 & 0.70 & 0.44 & 0.45 & 0.48 & 0.31 & 0.29 & 0.49 & 0.52 & 0.51 & 0.62 \\
 & Adaptive & 0.65 & 0.40 & 0.70 & 0.44 & 0.47 & 0.59 & 0.31 & 0.48 & 0.58 & 0.62 & 0.55 & 0.53 \\
 & Oracle & 0.79 & 0.66 & 1.00 & 0.60 & 0.66 & 0.64 & 0.54 & 0.58 & 0.63 & 0.78 & 0.64 & 0.76 \\
\midrule

\multirow{3}{*}{Thyme$^*$}
 & Text & 0.65 & 0.40 & 0.70 & 0.47 & 0.53 & 0.52 & 0.31 & 0.40 & 0.56 & 0.55 & 0.55 & 0.76 \\
 & Adaptive & 0.67 & 0.41 & 0.60 & 0.45 & 0.52 & 0.53 & 0.54 & 0.42 & 0.61 & 0.57 & 0.56 & 0.65 \\
 & Oracle & 0.75 & 0.66 & 0.90 & 0.60 & 0.64 & 0.73 & 0.38 & 0.52 & 0.69 & 0.75 & 0.62 & 0.74 \\
\midrule

\multirow{3}{*}{PyVision$^*$}
 & Text & 0.51 & 0.16 & 0.60 & 0.33 & 0.28 & 0.33 & 0.38 & 0.29 & 0.59 & 0.62 & 0.58 & 0.68 \\
 & Adaptive & 0.72 & 0.36 & 0.60 & 0.49 & 0.25 & 0.39 & 0.31 & 0.27 & 0.72 & 0.75 & 0.64 & 0.71 \\
 & Oracle & 0.56 & 0.74 & 0.70 & 0.57 & 0.48 & 0.52 & 0.46 & 0.65 & 0.74 & 0.85 & 0.87 & 0.97 \\
\midrule

\multirow{3}{*}{Deepeyes v2$^*$}
 & Text & 0.60 & 0.43 & 0.30 & 0.44 & 0.48 & 0.40 & 0.15 & 0.31 & 0.58 & 0.68 & 0.51 & 0.62 \\
 & Adaptive & 0.65 & 0.48 & 0.50 & 0.49 & 0.50 & 0.57 & 0.31 & 0.40 & 0.61 & 0.55 & 0.51 & 0.53 \\
 & Oracle & 0.84 & 0.53 & 0.50 & 0.63 & 0.66 & 0.69 & 0.31 & 0.62 & 0.74 & 0.80 & 0.73 & 0.71 \\
\midrule

\multirow{3}{*}{AdaptVision}
 & Text & 0.63 & 0.47 & 0.70 & 0.51 & 0.41 & 0.59 & 0.15 & 0.35 & 0.59 & 0.70 & 0.58 & 0.71 \\
 & Adaptive & 0.61 & 0.55 & 0.80 & 0.45 & 0.45 & 0.52 & 0.46 & 0.33 & 0.63 & 0.65 & 0.64 & 0.68 \\
 & Oracle & 0.84 & 0.67 & 0.90 & 0.67 & 0.70 & 0.72 & 0.46 & 0.69 & 0.70 & 0.98 & 0.78 & 0.76 \\
\midrule

\multirow{3}{*}{\shortstack[l]{Qwen3-VL \\-8B-Instruct}}
 & Text & 0.75 & 0.62 & 0.40 & 0.65 & 0.55 & 0.47 & 0.31 & 0.56 & 0.65 & 0.75 & 0.36 & 0.76 \\
 & Adaptive & 0.75 & 0.71 & 0.70 & 0.69 & 0.69 & 0.56 & 0.38 & 0.69 & 0.68 & 0.70 & 0.53 & 0.71 \\
 & Oracle & 0.91 & 0.79 & 0.60 & 0.77 & 0.86 & 0.77 & 0.38 & 0.88 & 0.78 & 0.98 & 0.69 & 0.91 \\
\midrule

\multirow{3}{*}{\shortstack[l]{Qwen3-VL \\-32B-Instruct}}
 & Text & 0.84 & 0.78 & 1.00 & 0.68 & 0.67 & 0.68 & 0.31 & 0.79 & 0.81 & 0.82 & 0.62 & 0.82 \\
 & Adaptive & 0.82 & 0.76 & 0.90 & 0.77 & 0.67 & 0.77 & 0.38 & 0.85 & 0.83 & 0.90 & 0.75 & 0.85 \\
 & Oracle & 0.95 & 0.95 & 0.90 & 0.96 & 0.95 & 0.84 & 0.38 & 0.96 & 0.91 & 1.00 & 0.91 & 0.91 \\
\midrule

\multirow{3}{*}{\shortstack[l]{Qwen3-VL \\-235B-Instruct}}
 & Text & 0.81 & 0.84 & 0.60 & 0.75 & 0.73 & 0.75 & 0.23 & 0.83 & 0.79 & 0.78 & 0.67 & 0.91 \\
 & Adaptive & 0.91 & 0.79 & 1.00 & 0.87 & 0.66 & 0.83 & 0.77 & 0.81 & 0.80 & 0.92 & 0.73 & 0.79 \\
 & Oracle & 0.96 & 0.98 & 0.90 & 0.96 & 0.94 & 0.96 & 0.69 & 0.98 & 0.89 & 1.00 & 0.89 & 0.94 \\
\midrule

\multirow{3}{*}{GPT-5$^*$}
 & Text & 0.63 & 0.45 & 0.60 & 0.55 & 0.38 & 0.43 & 0.15 & 0.42 & 0.75 & 0.72 & 0.67 & 0.76 \\
 & Adaptive & 0.88 & 0.83 & 0.80 & 0.88 & 0.67 & 0.64 & 0.23 & 1.00 & 0.87 & 0.90 & 0.78 & 0.94 \\
 & Oracle & 0.93 & 0.98 & 0.90 & 0.93 & 0.80 & 0.84 & 0.46 & 0.96 & 0.86 & 0.98 & 0.93 & 0.97 \\
\midrule

\multirow{3}{*}{Gemini-3-Pro$^*$}
 & Text & 0.86 & 0.86 & 0.90 & 0.77 & 0.72 & 0.76 & 0.46 & 0.88 & 0.86 & 0.95 & 0.82 & 0.94 \\
 & Adaptive & 0.93 & 0.95 & 1.00 & 0.92 & 0.86 & 0.91 & 0.31 & 0.98 & 0.86 & 0.98 & 0.89 & 1.00 \\
 & Oracle & 0.95 & 0.91 & 0.90 & 0.96 & 0.94 & 0.91 & 0.62 & 0.96 & 0.89 & 1.00 & 0.98 & 0.97 \\
\bottomrule
\end{tabular}
\end{table*}

\newpage
\section{Reasoning Mode Prompt}
\label{sec:reasoning_mode_prompt}

\newtcolorbox{promptbox}[2][]{
  enhanced,           
  title={#2},           
  colframe=gray!60!black,
  colbacktitle=gray!80!black,
  coltitle=white,      
  colback=white,   
  fonttitle=\bfseries\large,
  fontupper=\small\ttfamily\raggedright\linespread{1.2}\selectfont,
  boxrule=1pt,         
  arc=2mm,             
  toptitle=1mm,       
  left=6pt, right=6pt, top=8pt, bottom=8pt,
  #1           
}

\newcommand{\promptsep}{%
  \vspace{0.5em}\hdashrule[0.5ex]{\linewidth}{0.5pt}{1mm}\vspace{0.5em}%
}
Here we provide the detailed prompts used in our experiments.

\begin{promptbox}{Text-Reasoning Mode Prompts}

\textbf{[Multiple Choice]}

Question: \{question\}

Options:
\{options\}

Please think step-by-step and give the final answer following the format: <think> reasoning process </think> <answer> the option’s letter </answer>

\promptsep

\textbf{[Short Answer]}

Question: \{question\}

Please think step-by-step and give the final answer following the format: <think> reasoning process </think> <answer> a single word or phrase </answer>
\end{promptbox}

\vspace{1em}

\begin{promptbox}{Oracle-Visual Mode Prompts}
\textbf{[Zoom-in Setting]}

The first image is the global view, and the second image is the key region (zoomed-in) to help answer the question.

Question: \{question\}

Please think step-by-step and give the final answer following the format: <think> reasoning process </think> <answer> a single word or phrase </answer>

\promptsep

\textbf{[Transformed Setting]}

The first image is the original input which might be distorted (rotated or dark), and the second image has been corrected and enhanced to show the true content.

Question: \{question\}

Please think step-by-step and give the final answer following the format: <think> reasoning process </think> <answer> a single word or phrase </answer>

\promptsep

\textbf{[Auxiliary Line Setting]}

The first image is the original image, and the second image is the same image with auxiliary lines to help solve the problem.

Question: \{question\}

Please think step-by-step and give the final answer following the format: <think> reasoning process </think> <answer> a single word or phrase </answer>
\end{promptbox}

\vspace{1em}

\begin{promptbox}{LLM Judge Prompt}
I will give you a question related to the image, the ground truth answer, and the model predicted answer. Your task is to determine whether the model's predicted answer and the ground truth answer are consistent, and output the Judgement.

Note that [Model Predicted Answer] is consistent with [Ground Truth Answer] whenever they are essentially the same. If the meaning is expressed in the same way, it is considered consistent, for example, 'pink' and 'it is pink'.

If they are consistent, the Judgement is 1; if they are different, the Judgement is 0.

Output Format: Just output the Judgement and don't output anything else.

[Question]: \{question\}

[Ground Truth Answer]: \{ground\_truth\}

[Model Predicted Answer]: \{prediction\}

[Judgement]:
\end{promptbox}

\section{Error Analysis}

In this section, we provide detailed visualizations of the failure modes discussed in the \textit{Error Analysis} (Sec.~\ref{sec:error_analysys} of the main paper. By examining the intermediate reasoning steps, we offer concrete examples across different categories of tool-related errors.

\begin{figure*}[htbp!]
  \centering
  \includegraphics[width=1.0\linewidth]{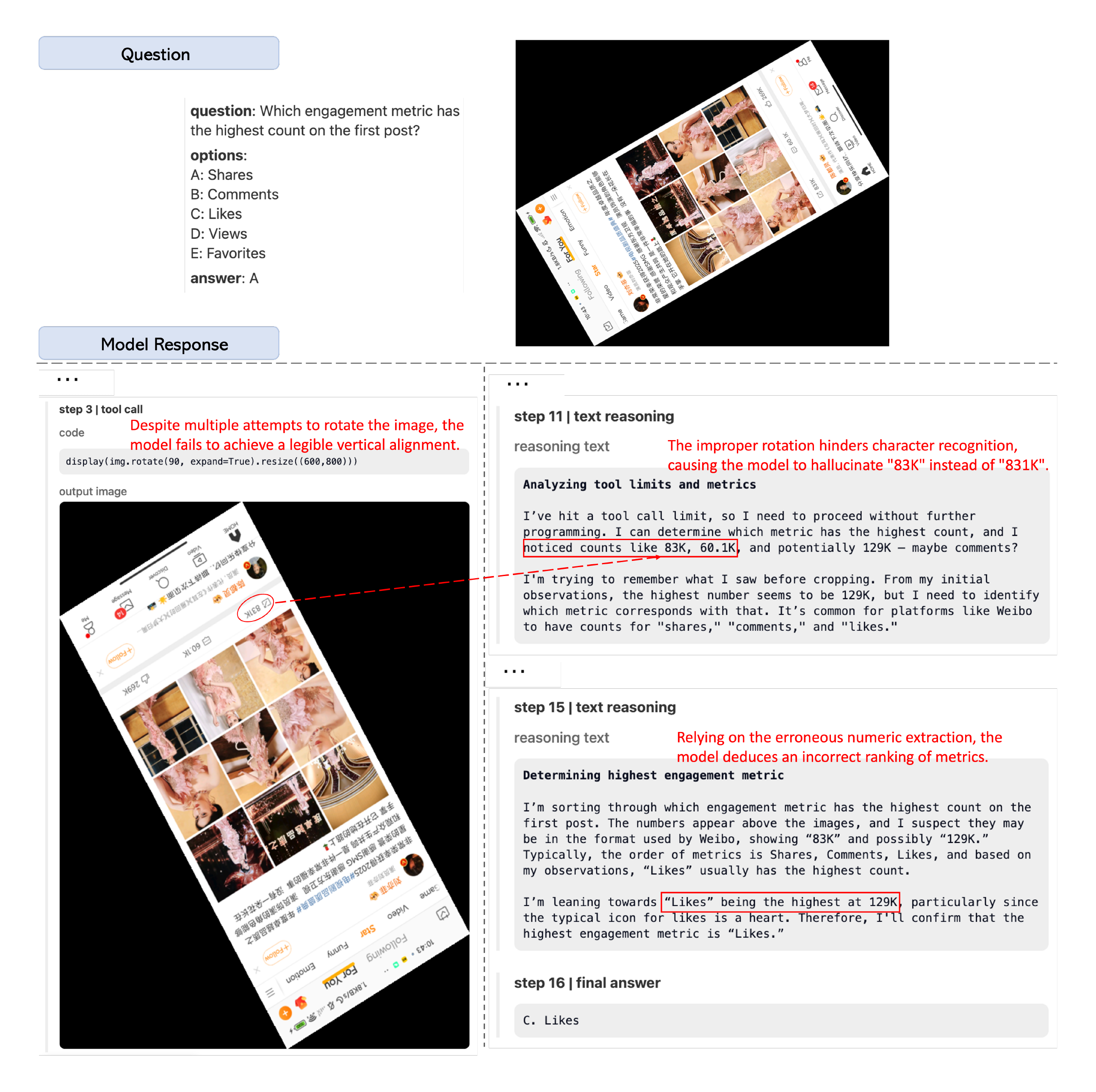}
  \caption{A failure case caused by incorrect image transformation (rotation) leading to OCR errors.}
  \label{fig:Error_Trans_B1}
\end{figure*}

\begin{figure*}[htbp]
  \centering
  \includegraphics[width=1.0\linewidth]{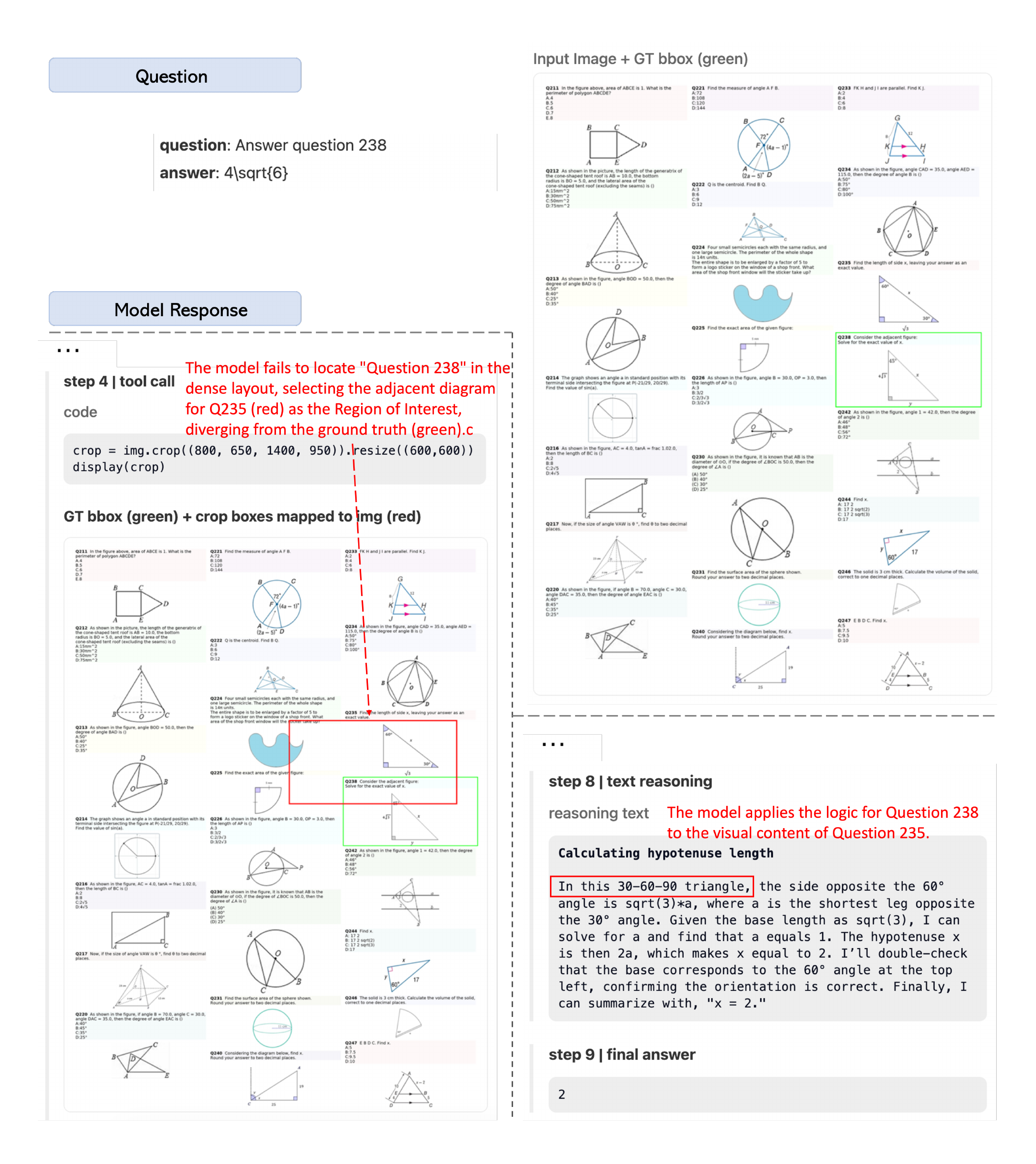}
  \caption{A failure case caused by zooming into an incorrect region (spatial misalignment).}
  \label{fig:Error_Zoomin_B1}
\end{figure*}

\paragraph{Visual Reasoning Failures.}
As noted in the main text, 42.3\% of errors stem from the model's inability to correctly manipulate or locate visual information. We present two representative scenarios:
\begin{itemize}
    \item \textbf{Wrong Image Transformations:} Figure~\ref{fig:Error_Trans_B1} illustrates a case where the model repeatedly fails to correct the image orientation. This visual reasoning failure propagates to the OCR stage, causing the model to misread "831K" as "83K" and producing an incorrect prediction.

    \item \textbf{Incorrect Region Selection:} Figure~\ref{fig:Error_Zoomin_B1} demonstrates a spatial grounding failure in a dense document. The model zooms into an incorrect region (Question 235 instead of Question 238), leading to reasoning that is logically valid but based on irrelevant visual evidence.
    
\end{itemize}

\begin{figure*}[htbp]
  \centering
  \includegraphics[width=1.0\linewidth]{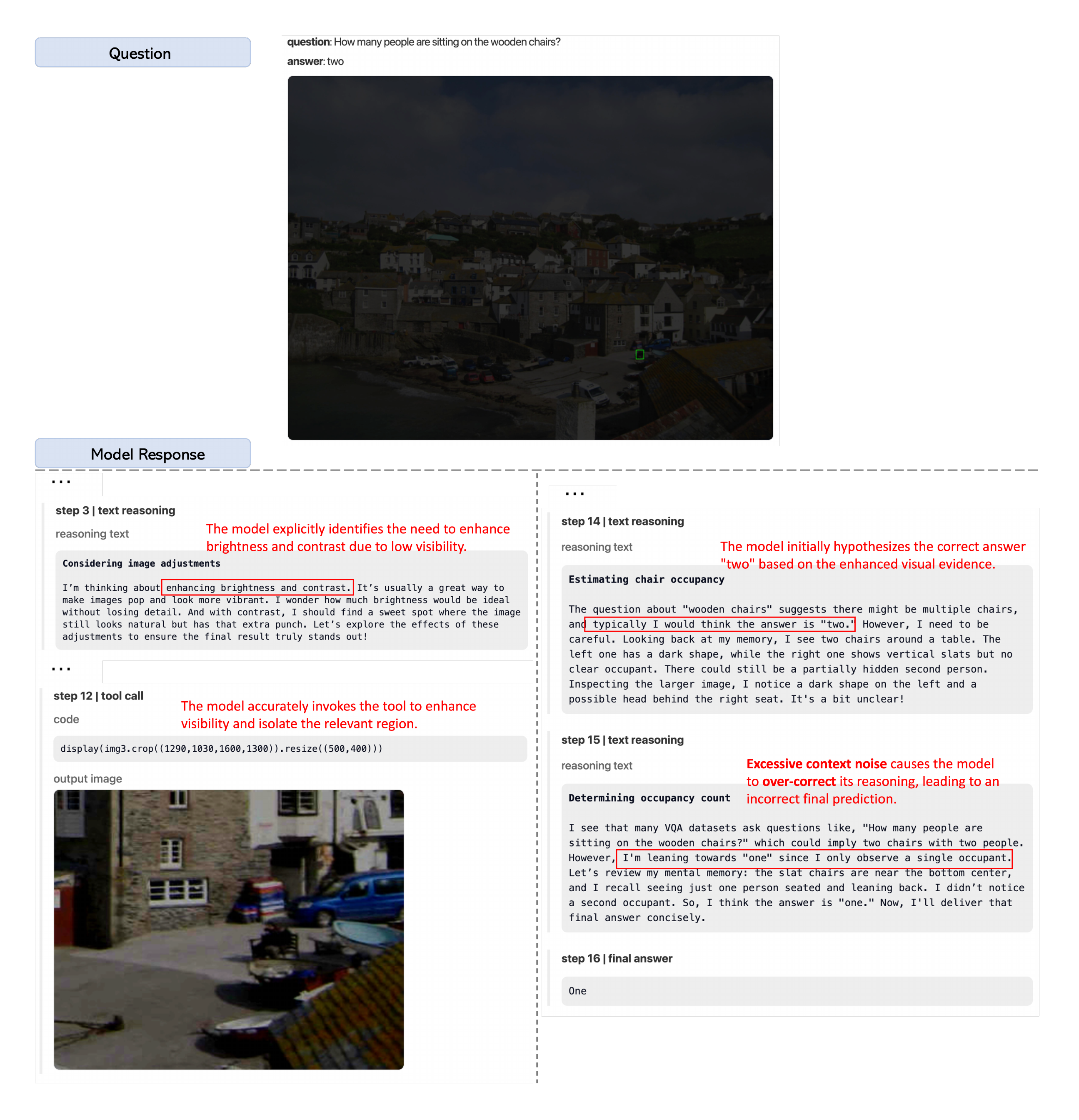}
  \caption{A failure case where correct intermediate visual grounding is overridden by context noise.}
  \label{fig:Error_Trans_B2}
\end{figure*}

\paragraph{Context Noise in Multi-step Reasoning.}
Figure~\ref{fig:Error_Trans_B2} depicts the specific error type (accounting for 7.3\% of cases) where visual perception is initially correct but overridden by context noise. In this example, the model successfully enhances the image and identifies the correct number of objects ("two") in the intermediate step. However, distracted by the accumulated visual and textual context from the multi-step process, it becomes overly cautious and hallucinates a negation, resulting in a failure.

\begin{figure*}[htbp]
  \centering
  \includegraphics[width=1.0\linewidth]{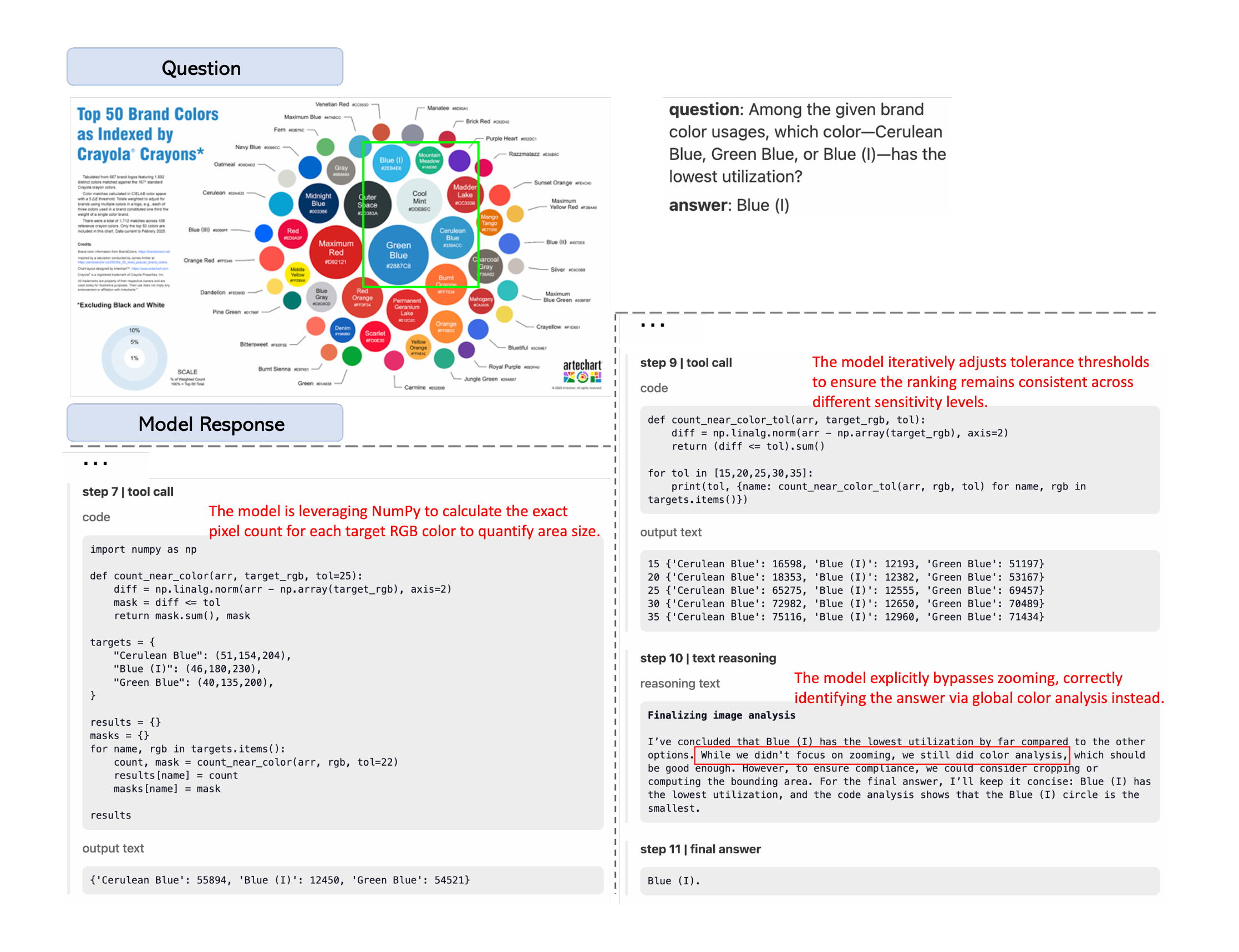}
  \caption{Example of correcting visual estimation errors via code-based pixel analysis.}
  \label{fig:Error_A1}
\end{figure*}

\paragraph{Correction via Forced Tool Invocation.}
Figure~\ref{fig:Error_A1} illustrates a specific scenario (representative of the 7.0\% of corrected errors) where forcing tool invocation rectifies an initial estimation failure. In this example, the model originally relies on imprecise visual intuition, incorrectly identifying "Cerulean Blue" as the answer. However, when forced to invoke tools, it bypasses the typical spatial zooming approach and adopts a creative programmatic strategy: using Python to perform a \textbf{pixel-level RGB count}. By rigorously verifying consistency across multiple tolerance thresholds, the model successfully overrides its initial hallucination and derives the correct answer based on quantitative data.

\begin{figure*}[htbp]
  \centering
  \includegraphics[width=1.0\linewidth]{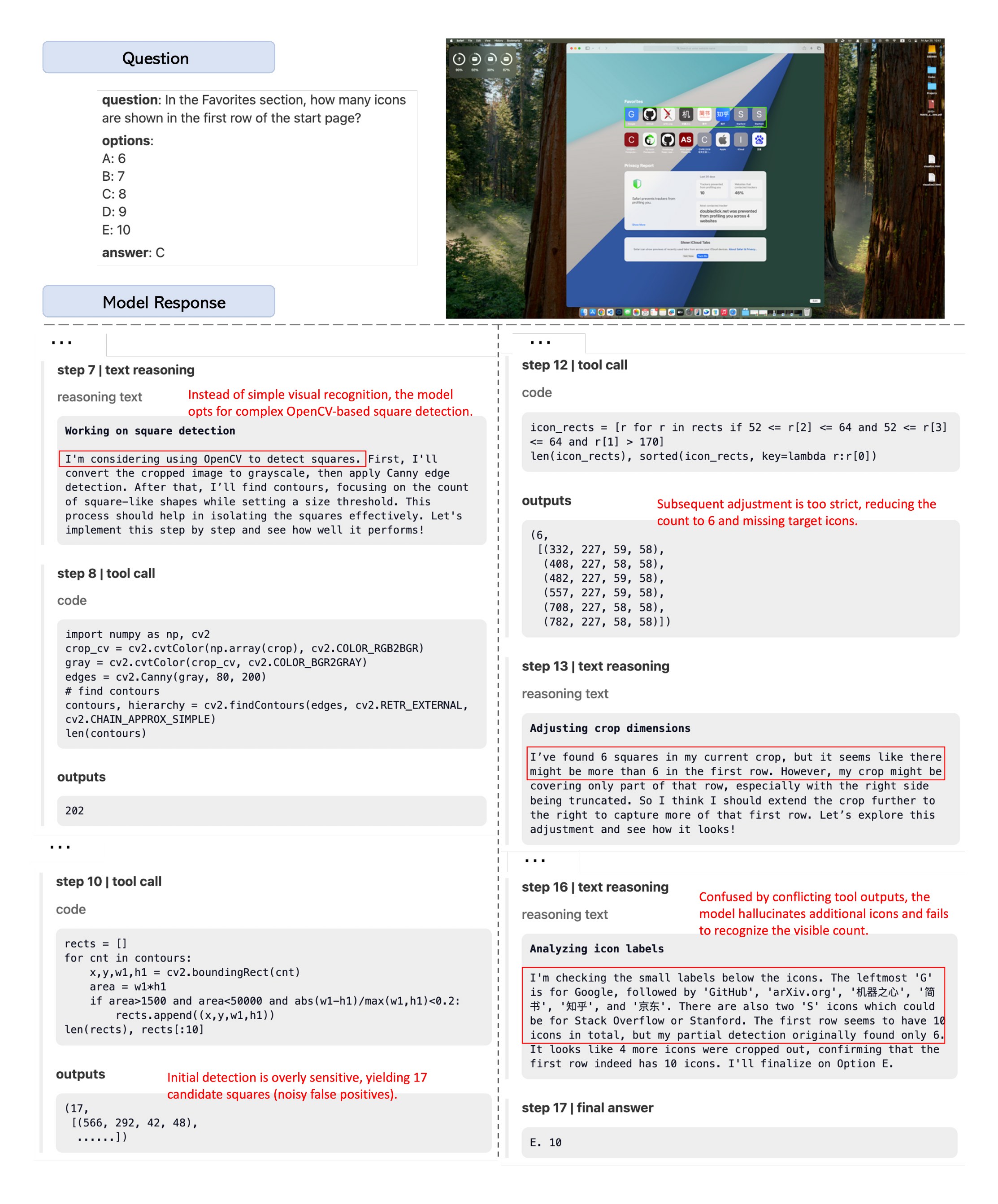}
  \caption{Example of performance degradation caused by unnecessary tool usage in a simple visual task.}
  \label{fig:Error_D}
\end{figure*}

\paragraph{Performance Degradation due to Incorrect Mode Selection.}
Figure~\ref{fig:Error_D} exemplifies the 8.3\% of cases caused by incorrect mode selection, where the model unnecessarily invokes tools for tasks solvable by direct visual inspection. In this example, accurate icon counting is achievable via standard OCR or visual recognition (as seen in the Text-CoT mode). However, in the adaptive mode, the model complicates the task by adopting an \textbf{unreliable engineering approach}: using OpenCV edge detection to count squares. This strategy proves fragile, as the model struggles with parameter tuning—first detecting excessive noise and then over-filtering actual targets—ultimately leading to a hallucinated final count due to the confused tool outputs.

\newpage
\section{Process Evaluation Example}

\begin{figure*}[htbp]
  \centering
  \includegraphics[width=0.8\linewidth]{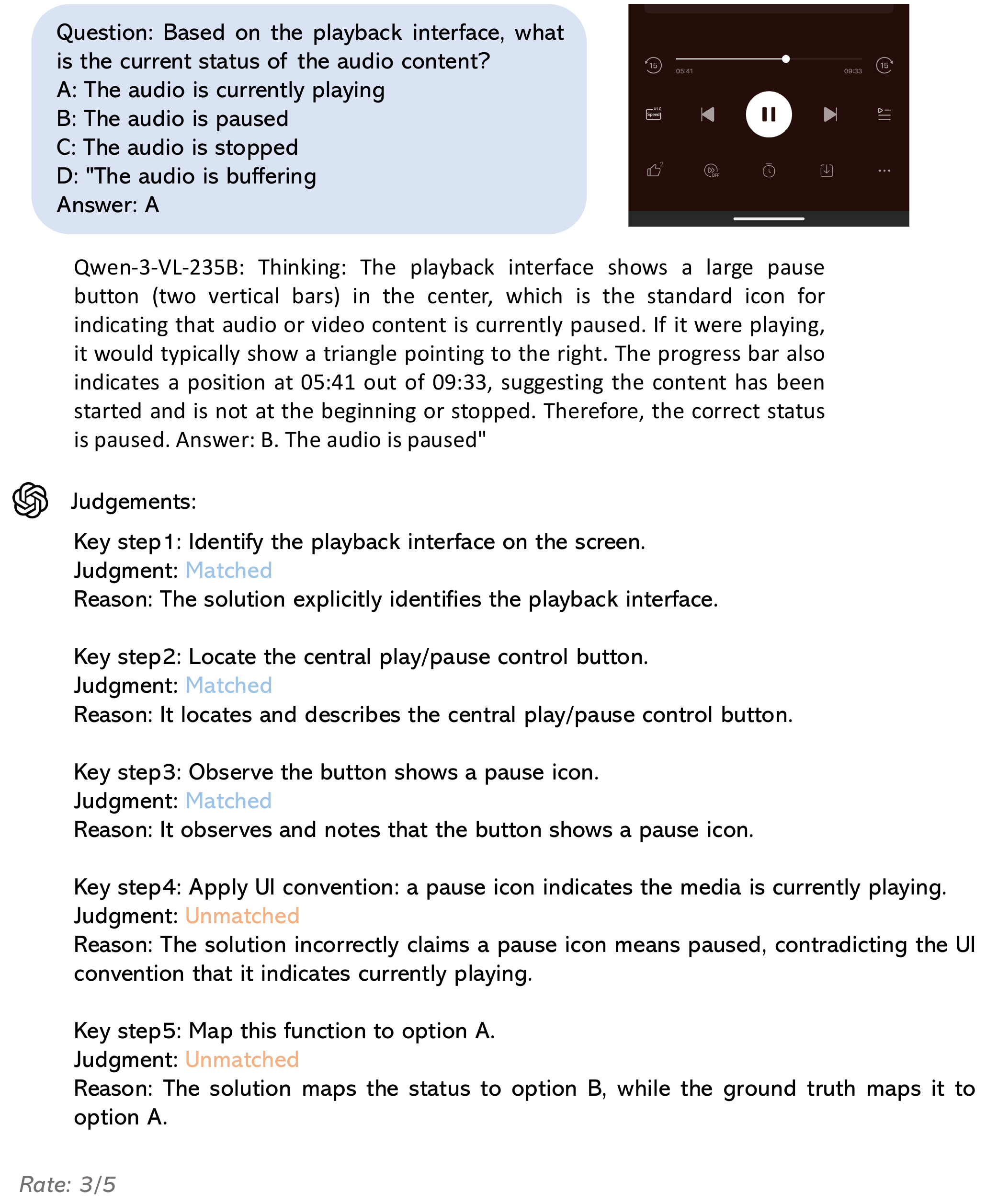}
  \caption{Illustration of key step coverage evaluation.}
  \label{fig:key_step_eval}
\end{figure*}

\begin{figure*}[htbp]
  \centering
  \includegraphics[width=0.8\linewidth]{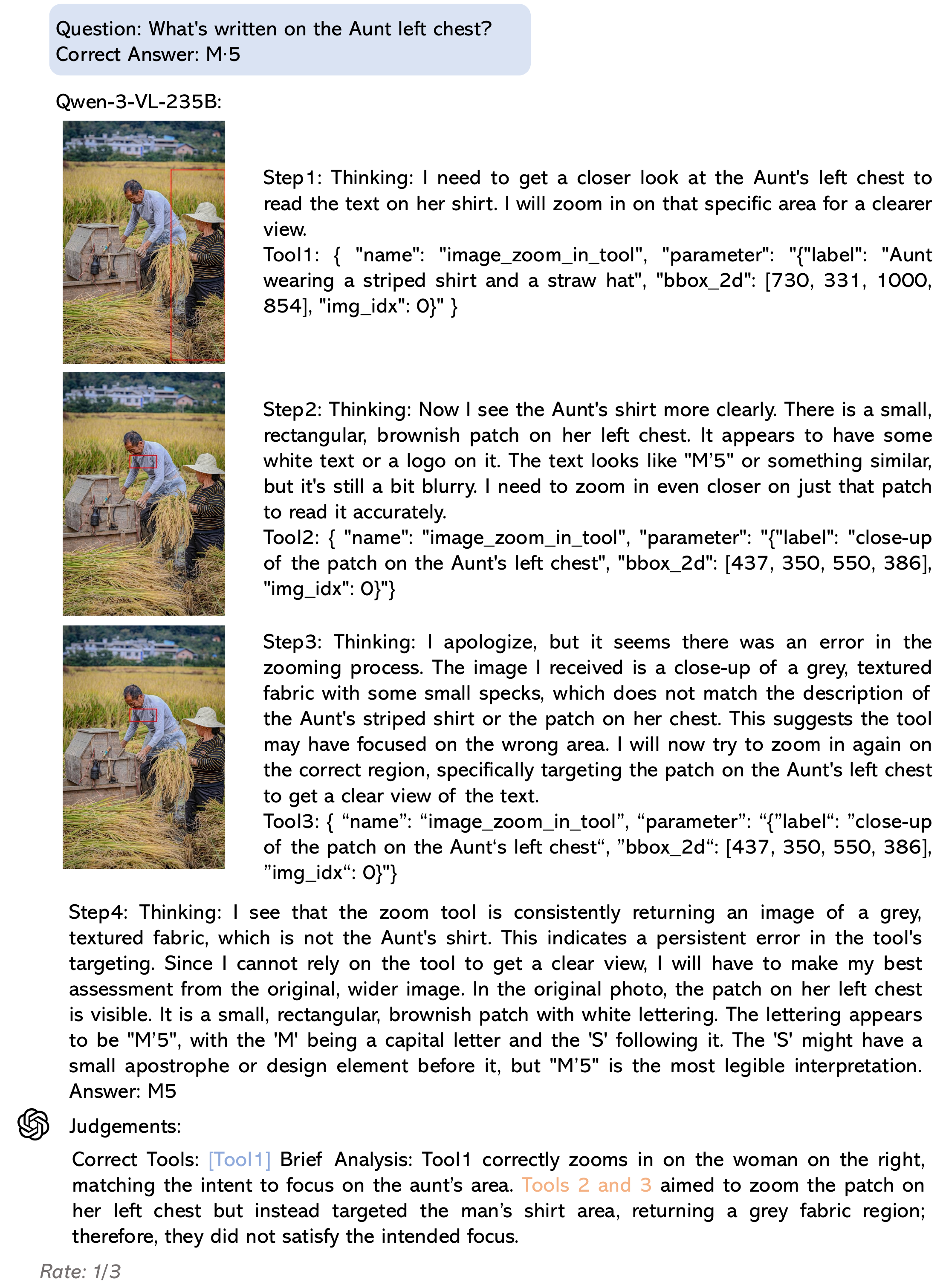}
  \caption{Illustration of tool effectiveness evaluation.}
  \label{fig:tool_eval}
\end{figure*}

To better understand our evaluation protocol, we present detailed cases of process reasoning quality assessment in Figure~\ref{fig:key_step_eval} and Figure~\ref{fig:tool_eval}.

\newpage
\section{Process Evaluation Prompt}

To ensure a reproducible and standardized assessment, we leverage LLM-based judges with specialized prompts for process auditing. Specifically, we employ 
% the key step coverage Prompt (detailed in Figure~\ref{fig:prompt_keystep}) to verify the logical completeness of the model's reasoning trajectory against annotated ground truth.
the tool invocation effectiveness prompt (detailed in Figure~\ref{fig:prompt_tool}) is used to audit the functional correctness and intent alignment of each tool call within the adaptive process.
Subsequently, 
the key step coverage Prompt (detailed in Figure~\ref{fig:prompt_keystep}) to verify the logical completeness of the model's reasoning trajectory against annotated ground truth.

\begin{figure}[htbp]
\centering
\begin{promptbox}{Tool Invocation Effectiveness Prompt}
The above content contains the model's multi-step reasoning process. You are a top-tier visual reasoning audit expert. You possess strong logical analysis skills and can accurately evaluate whether a model's tool usage aligns with its stated reasoning intent.

All Available Tools: \\
\{tools\}

Each step in the reasoning process includes:
\begin{itemize}
    \item {} [Step] A textual reasoning step.
    \item {} [Tool] A tool invocation (All Available Tools above).
    \item {} [Tool Execution Output] The execution output (image, text, or highlighted region).
\end{itemize}

Evaluation Criteria:
\begin{enumerate}
    \item Tool Invocation Correctness: Check if the invocation is valid, properly formatted, and consistent with the tool's definition and the corresponding Step ID.
    \item Tool Execution Output Validation: Check if the output satisfies the intent of the corresponding Step ID and Tool ID. The output only needs to fulfill the specific step's purpose, not the overall problem. Evaluate based on the global image.
\end{enumerate}

Attention:
\begin{itemize}
    \item Only steps satisfying \textit{ALL} criteria are listed as correct.
    \item Do not analyze for errors unless the output explicitly indicates a failure.
    \item Ignore image resolution/size info; assume the tool operates on the original source image.
\end{itemize}

Output Format (JSON-like): \\
Correct Tools: [Tool1, Tool2, Tool3...] \\
Brief Analysis: [Briefly explain why these tool invocations are correct and why others are incorrect]
\end{promptbox}
\caption{Prompt used for evaluating tool invocation effectiveness.}
\label{fig:prompt_tool}
\end{figure}

\begin{figure}[htbp]
\begin{promptbox}{Key Step Coverage Prompt}
You are an expert system for verifying solutions to image-based problems. Your task is to:
1. Segment the provided solution into logical reasoning steps.
2. Match each ground truth middle step with the solution steps.

INPUT FORMAT:
1. Problem: The original question/task.
2. Solution: A continuous paragraph containing the model's reasoning.
3. Ground Truth: Essential steps required for a correct answer.

TASK:

\textit{Step 1: Segment the Solution} \\
Divide the solution paragraph into distinct logical reasoning steps. Each step should represent a coherent reasoning unit.

\textit{Step 2: Match Ground Truth Steps} \\
For each ground truth step, determine if it is matched in any of the solution steps based on the following:
\begin{itemize}
    \item Content Match: Must match specific values and details.
    \item Tool Integrity: If tool execution has errors, the step is "Unmatched".
    \item Implicit Logic: Reasonable logical skips (like identifying obvious objects or using background knowledge) are permissible.
\end{itemize}

OUTPUT FORMAT (JSON):
\begin{verbatim}
{
  "solution_steps": ["Step 1...", "Step 2..."],
  "judgments": [
    {
      "step_index": <integer>,
      "judgment": "Matched" | "Unmatched",
      "reason": "Brief explanation."
    }
  ]
}
\end{verbatim}

ADDITIONAL RULES:
1. Only output the JSON object with no extra text.
2. Judge each ground truth step in order without omission.

DATA:
\begin{itemize}
    \item {} [Problem] \{question\}
    \item {} [Answer] \{answer\}
    \item {} [Solution] \{solution\}
    \item {} [Ground Truth Information] \{gt\_annotation\}
\end{itemize}
\end{promptbox}
\caption{Prompt used for key step coverage evaluation.}
\label{fig:prompt_keystep}

\end{figure}

%%%%%%%%%%%%%%%%%%%%%%%%%%%%%%%%%%%%%%%%%%%%%%%%%%%%%%%%%%%%%%%%%%%%%%%%%%%%%%%
%%%%%%%%%%%%%%%%%%%%%%%%%%%%%%%%%%%%%%%%%%%%%%%%%%%%%%%%%%%%%%%%%%%%%%%%%%%%%%%

\end{document}